\documentclass[10pt,twocolumn,letterpaper]{article}

\usepackage{amsmath}
\usepackage{amssymb}
\usepackage{pifont}

\usepackage{3dv}
\usepackage{times}
\usepackage{epsfig}
\usepackage{graphicx}

\usepackage{booktabs}
\usepackage{mwe}
\usepackage{colortbl}
\usepackage{multirow}
\usepackage{color, soul}
\usepackage{arydshln}

\usepackage[utf8]{inputenc}
\usepackage{hhline}
\usepackage{colortbl}
\usepackage{multicol}

\usepackage{enumitem}
\setlist{nosep}

\usepackage{floatrow}
\usepackage{siunitx}

\usepackage{subcaption}
\usepackage[font=small,labelfont=bf]{caption}

\usepackage[ruled,vlined]{algorithm2e}
\usepackage{etoolbox}
\AtBeginEnvironment{algorithm}{\SetArgSty{textrm}}
\SetKwFor{For}{for (}{) $\lbrace$}{$\rbrace$}
\makeatletter
\patchcmd{\@algocf@start}
  {-1.5em}
  {0pt}
  {}{}
\makeatother

\SetCommentSty{mycommfont}

\usepackage[export]{adjustbox}

\usepackage{dsfont}

\newif\ifshowedits

\newcommand{\addeditor}[3]{%
  \definecolor{#1color}{rgb}{#3}
  \expandafter\newcommand\csname #1\endcsname[1]{%
  \ifshowedits
    {\color{#1color} ##1}%
  \else
    {##1}%
  \fi
  }%
  \expandafter\newcommand\csname #1rmk\endcsname[1]{%
  \ifshowedits
    {\color{#1color} {\bf [#2: ##1]}}
  \fi
  }%
  \expandafter\newcommand\csname #1rpl\endcsname[2]{%
  \ifshowedits
    {\color{#1color} ##1 \sout{##2}}
  \else
    {##1}
  \fi
  }%
}

\newcommand{\mycomment}[1]{}

\newcommand{\calL}{{\cal L}}

\newcommand{\bR}{{\bf R}}

\newcommand{\bT}{{\bf T}}

\newcommand{\IR}{{\mathds{R}}}

\DeclareFontFamily{U}{mathx}{\hyphenchar\font45}
\DeclareFontShape{U}{mathx}{m}{n}{
      <5> <6> <7> <8> <9> <10>
      <10.95> <12> <14.4> <17.28> <20.74> <24.88>
      mathx10
      }{}
\DeclareSymbolFont{mathx}{U}{mathx}{m}{n}
\DeclareFontSubstitution{U}{mathx}{m}{n}
\DeclareMathAccent{\widebar}{0}{mathx}{"73}

\newcommand{\crop}{\text{crop}}

\newcommand{\bomega}{\boldsymbol{\omega}}

\newcommand{\feat}{\mathbf{f}}
\newcommand{\encoder}{\Phi}
\newcommand{\regressor}{\Psi}
\newcommand{\transformer}{\mathcal{T}}

\newcolumntype{d}[1]{D{.}{.}{#1}}
\newcolumntype{R}{@{\extracolsep{3cm}}r@{\extracolsep{0pt}}}

\newcommand*\samethanks[1][\value{footnote}]{\footnotemark[#1]}
\showeditsfalse

\usepackage[pagebackref=true,breaklinks=true,letterpaper=true,colorlinks,bookmarks=false]{hyperref}

\threedvfinalcopy 

\def\threedvPaperID{162} 

\ifthreedvfinal\fi
\begin{document}
\title{PIZZA: A Powerful Image-only Zero-Shot Zero-CAD Approach \\ to 6~DoF Tracking
}
\author{
	\thanks{The first two authors contributed equally.}~ Van Nguyen Nguyen, \quad \samethanks~ Yuming Du \quad Yang Xiao \quad Micha\"el Ramamonjisoa \quad Vincent Lepetit \\
	LIGM, Ecole des Ponts, Univ Gustave Eiffel, CNRS, France\\
	{\tt\small\{firstname.lastname\}@enpc.fr }
}
\vspace{-8mm}

\maketitle
\thispagestyle{plain}
\pagestyle{plain}
\begin{abstract}
Estimating the relative pose of a new object without prior knowledge is a hard problem, while it is an ability very much needed in robotics and Augmented Reality. We present a method for tracking the 6D motion of objects in RGB video sequences when neither the training images nor the 3D geometry of the objects are available. In contrast to previous works, our method can therefore consider unknown objects in open world instantly, without requiring any prior information or a specific training phase. We consider two architectures, one based on two frames, and the other relying on a Transformer Encoder, which can exploit an arbitrary number of past frames. We train our architectures using only synthetic renderings with domain randomization. Our results on challenging datasets are on par with previous works that require much more information~(training images of the target objects, 3D models, and/or depth data). Our source code is available at \href{https://github.com/nv-nguyen/pizza}{https://github.com/nv-nguyen/pizza}.
\end{abstract}

\vspace{-6mm}

\section{Introduction}
\label{sec:introduction}
\vspace{-1mm}
To evolve in open worlds and interact with unknown objects, autonomous systems will require the ability to track the 6D pose of objects without any prior information, in an extreme case, the system is required to track the 6D pose of the object given only one color image. However, current approaches to 6D pose estimation still have requirements preventing this. They may require a large number of images annotated with the object's 6D poses, which are cumbersome to acquire~\cite{kehl-iccv17-ssdd,rad-iccv17-bb}. Even when the training images are synthetic, a 3D model and/or a long training phase may be needed~\cite{labbe2020,park-eccv20-few}. 

Recently, a few methods have attempted to predict the 6D pose of objects without specifically training for these objects, by exploring meta-learning~\cite{tseng2019few}, correspondences between the image and a 3D model of the object~\cite{Pitteri20203DOD} or template matching~\cite{nguyen2022templates,Shugurov_2022_CVPR}. These methods, however, still require the object's 3D model, which is not likely to be available in open worlds. It is also possible to train a model on many examples from the same object category~\cite{grabner2019LFD,Wang_2019_NOCS}. Then, the model can generalize to new objects without 3D model nor a retraining phase, but only for objects belonging to the known categories.
\begin{figure}[t]
\newlength{\teaserheight}
\setlength\teaserheight{1.7cm}
\begin{center}
\begin{tabular}{cccc}
 \includegraphics[height=\teaserheight]{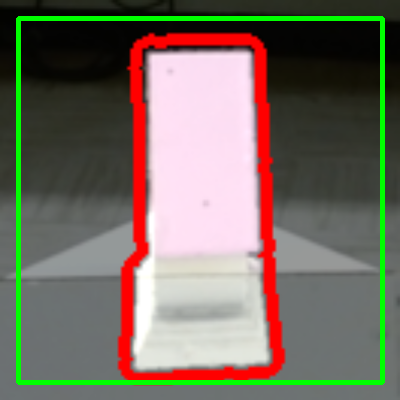} &
 \includegraphics[ height=\teaserheight]{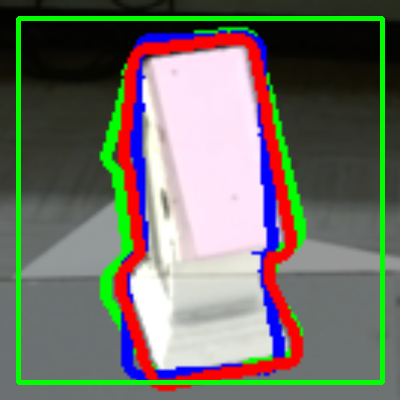} &
 \includegraphics[ height=\teaserheight]{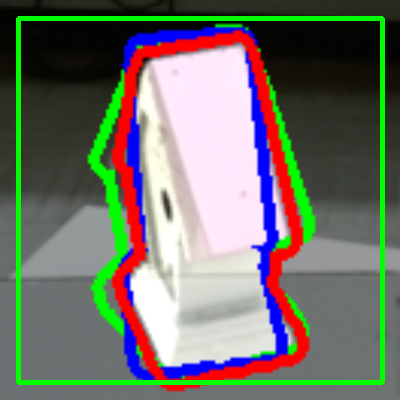} &
 \includegraphics[ height=\teaserheight]{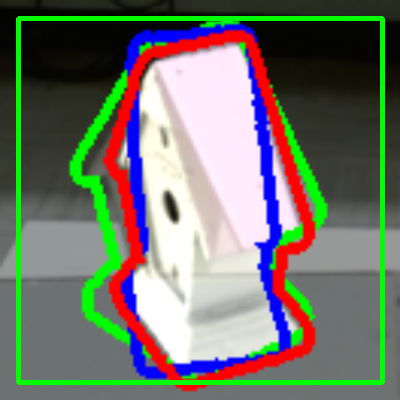}\\
 \includegraphics[height=\teaserheight]{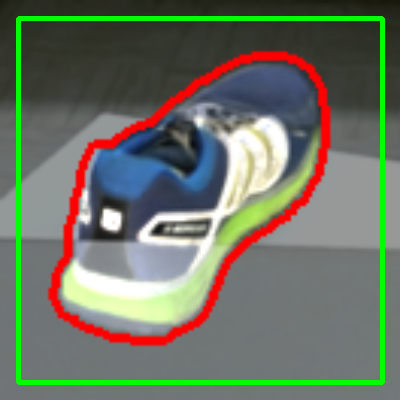} &
 \includegraphics[ height=\teaserheight]{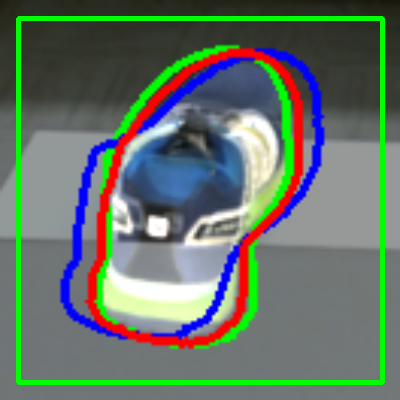} &
 \includegraphics[ height=\teaserheight]{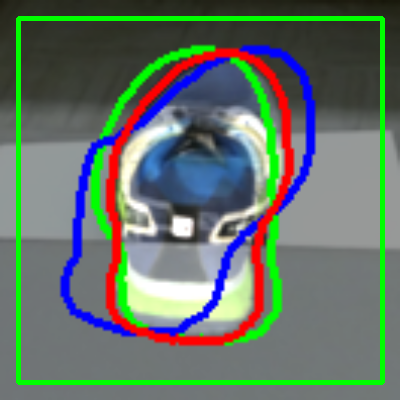} &
 \includegraphics[ height=\teaserheight]{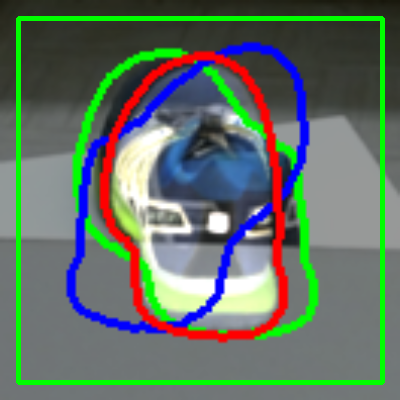}
\end{tabular}\\
\begin{tabular}{c}
\includegraphics[ height=0.22\teaserheight]{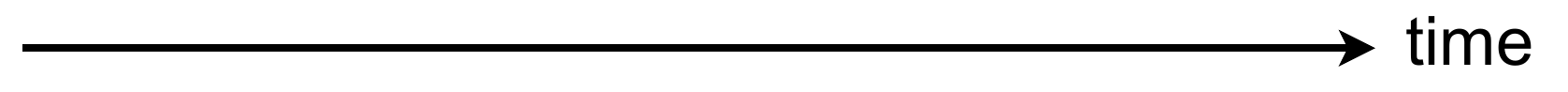}
\end{tabular}
\end{center}
\vspace{-8mm}
\caption{Given a 2D bounding box to select the object to track, our method recovers its 6D trajectory (3D translations and 3D rotations) for the next frames, without knowledge of the object geometry and without any training specific to the object. We use the ground truth 6D pose in the first frame and the 3D model \emph{for visualization only}: The 3D model is rendered using  6D poses computed from the pose in the first frame and the trajectory we recover. For each row, the left image is the first frame of the sequence, which contains the given region of interest in green. The \textbf{\color{blue}blue} and \textbf{\color{red}red} contours show the results of our two proposed methods, while the \textbf{\color{green}green} contours delineate the ground truth object mask.}
\label{fig:teaser}
\end{figure}

Recent methods such as LatentFusion~\cite{park2020latent}, Gen6D~\cite{liu2022gen6d}, OnePose~\cite{sun2022onepose} address the problem of not having the 3D models for the target objects by first acquiring and registering reference images taken from multiple viewpoints. LatentFusion and OnePose also require a training phase. BundleTrack~\cite{wen2021bundletrack} does not require a 3D model \emph{a priori} but relies on RGB-D images as input, which reveal the geometry of the object and thus simplify the problem. Therefore, being able to estimate the 6D pose of unkown objects from only RGB images remains an unsolved problem.

As shown in Fig.~\ref{fig:teaser}, in this paper, we demonstrate that it is possible to track the 6D pose of an object when no training images nor a 3D CAD model nor depth images are available for this object. Since we do not have access to the object's 3D model nor its coordinate frame, only a relative pose up to a scale factor can be estimated.  We therefore aim at predicting the 6D poses of unknown objects, not seen during training nor belonging to some specific category, relative to the first frame.

For our approach, which we call PIZZA for \textbf{P}owerful \textbf{I}mage-only \textbf{Z}ero-Shot \textbf{Z}ero-CAD \textbf{A}pproach, we rely on a Transformer encoder, which allows us to exploit an arbitrary number of past frames, in contrast with standard 3D tracking methods that typically rely on the previous frame only. Our Transformer-based method outputs the consecutive relative 6D poses between consecutive frames, which are then used recursively along with the initial object pose to provide the object 6D pose at each frame. As our experiments show, being able to exploit multiple frames reduces drift significantly compared to the standard frame-by-frame tracking framework.

One of the insights we had from our work is that the background has a strong influence on the accuracy of the predicted motions:  In our early experiments, our motion prediction appeared to struggle on unseen objects over a cluttered background. This is in contrast with 6D pose prediction for seen objects, which can be made robust simply by training on images with cluttered background. However, training our method using similar complex backgrounds did not prove to be sufficient for unseen objects.

To alleviate the effect of complex background, we leverage recent progress in 2D unseen instance segmentation: We noticed that when trained on many examples, object segmentation networks generalize surprisingly well on unseen objects, even objects from unseen categories. We rely on the recent work \cite{Du20211stPS} to get the objects's masks in the input videos and discard the background. Even when the masks are noisy, this improves motion prediction significantly.

As mentioned above, the object motion can be retrieved only up to a scale factor. To guarantee the motion predictions are consistent across frames, special care is required: In particular, the magnitudes of the translations should  be all defined up to the \emph{same} scale factor. We propose a suitable representation of the motion at the beginning of Section~\ref{sec:method}.

We first validate our method on unseen categories with synthetic data created from CAD models of ModelNet~\cite{Wu_2015_CVPR} as done in \cite{li-eccv18-deepim,Sundermeyer2020MultiPathLF,park2020latent} and synthetic videos created from CAD models of ShapeNet~\cite{chang2015shapenet}. Then, we train our approach on a dataset made only of synthetic data created from rendering CAD models from BOP challenges \cite{hodan2018bop} and test on real-world datasets Moped~\cite{park2020latent} and Laval \cite{garon2018framework}. We compare our methods with the state-of-the-art methods on deep 6D tracking methods: DeepIM~\cite{li-eccv18-deepim}, MultiPath Learning~\cite{Sundermeyer2020MultiPathLF}, LatentFusion~\cite{park2020latent}, MoveIt~\cite{busam2020moveit} and Laval~\cite{garon2018framework} which rely either 3D object models, or depth maps.

Thanks to domain randomization and realistic rendering, we achieve good generalization from synthetic to real data. Our method generalization capabilities are then two-fold, as it generalizes across domains, but also across novel object categories. We validate this by training only using synthetic data, and by making sure training and test objects belong to very dissimilar categories.

We believe our approach opens a new line of research on 6~DoF object tracking in open worlds, since to the best of our knowledge, it is the first method performing on RGB images that generalizes to unseen object categories without requiring any 3D information at train or test time such as depth maps or CAD models, nor pre-acquired reference images of the objects.
\vspace{-2mm}

\section{Related Work}
\label{sec:relatedwork}
\vspace{-2mm}
In this section, we first give an overview of 6 DoF object pose estimation, from single-view estimation methods to tracking methods that use multiple images for inference. We then briefly discuss unseen object segmentation and 2D tracking, as this aspect is important in our approach.

\subsection{6~DoF Object Pose Estimation}
\vspace{-1.5mm}
\paragraph{Monocular 6D pose estimation.} With the construction of various benchmarks for 6D object pose estimation~\cite{Hinterstoier2012ModelBT,Hodan2017TLESSAR,kaskman2019homebreweddb,rennie2016dataset,denninger2019blenderproc}, deep learning-based approaches have become dominant in this field. These methods can be roughly divided into three types of approaches: direct regression of the object pose~\cite{kehl-iccv17-ssdd,xiang2018posecnn}, template-based matching~\cite{Hinterstoier2012ModelBT,Sundermeyer2018Implicit3O} that encodes images in latent spaces and compares them against a dictionary of predefined viewpoints, and keypoint prediction for predicting the 6D pose~\cite{rad-iccv17-bb,Tekin2017RealTimeSS,peng2019pvnet,Hu19,Zhou2019Rotation6D} or dense 2D-3D correspondences~\cite{Li2019CDPNCD,Wang_2019_NOCS,Zakharov2019DPOD6P,Park2019Pix2PosePC} as in earlier works~\cite{Taylor12,Brachmann2014Learning6O}. 

Most existing learning-based methods are designed for known objects; very few works report performances on unseen objects by using object descriptors~\cite{Wohlhart2015LearningDF,Balntas2017PoseGR,Sundermeyer2020MultiPathLF}, generic 2D-3D correspondences shared for all objects~\cite{Pitteri20203DOD}, or latent 3D representations~\cite{park2020latent}. Recently, Gen6D~\cite{liu2022gen6d} and  OnePose~\cite{sun2022onepose} proposed an extension of LatentFusion~\cite{park2020latent} with only RGB annotated reference images for unseen pose estimation. These methods still rely on a 3D model of the target object or on reference images of the object captured under multiple viewpoints; category-level methods can generalize to unseen objects of the known categories without knowing the exact 3D model~\cite{grabner2019LFD,Wang_2019_NOCS,chen2020category,Manhardt2020CPSIC}. While Gen6D and  OnePose do not require a 3D model of the object, they still require a sequence of images taken from different views of the target object as reference images. Capturing the reference images still require human expertise and some time for capture and registration. 

\vspace{-4mm}
\paragraph{Temporal 6D pose tracking.} Instead of relying on a single image for absolute pose estimation, tracking methods exploit temporal information. While earlier methods were prone to fail in the presence of heavy occlusion and clutter~\cite{Kwon2007ParticleFO,choi13iros_rgbdtracking,Aldoma2013MultimodalCI}, data-driven methods have been proposed to learn more robust features by using Random Forests~\cite{Krull20146DOFMB,Tan2014MultiforestTA,Tan2015AVL}. This problem has been recently formulated under a deep learning framework, where a network is trained to regress the pose difference between image pairs extracted from RGB~\cite{deng2019pose,busam2020moveit} or RGB-D videos~\cite{garon2017deep6D,garon2018framework,Zhou2018DeepTAMDT,wang20196-pack}. 
\cite{garon2018framework,busam2020moveit, wen2021bundletrack} show that this network can be applied to new objects but still require 3D information in the form of a CAD model or depth images.

In contrast to aforementioned deep 6D tracking methods that rely on either 3D object models, or depth maps, we introduce a generic RGB-based tracking method without 3D models, which can be trained on synthetic images only. To the best of our knowledge, we are the first to report object pose tracking performance on unseen objects, without using 3D object models or depth maps at test time.

\subsection{2D Object Segmentation for Unseen Categories}

Object segmentation methods first considered a ``closed-world'' setting,  where the training and test sets share the same object classes. 
However, several authors, for example \cite{du2021learning}, noticed that a class-agnostic instance segmentation network is capable of generalizing to objects from unseen classes. The Unsupervised Video Objects~(UVO) dataset was recently introduced by \cite{wang2021unidentified} to rigorously benchmark the performance of different methods for segmentation of unseen classes. In this work, we selected  \cite{Du20211stPS} to detect and segment the objects, because it is currently the best performing method on UVO. \cite{Du20211stPS} relies on a two-stage network: The first stage consists of a class-agnostic object detector for proposal generation. The content in the proposals are fed to a foreground-background segmentation network to generate masks. This segmentation network relies on UPerNet~\cite{xiao2018unified} with Swin-L~\cite{liu2021swin} as backbone. It extracts features at multiple scales to predict robust and accurate masks.

\vspace{-3mm}

\section{Our 6D Tracking Method}
\label{sec:method}
Given an image stream capturing a moving object and its 3D pose in the first frame, we want to predict the 6D pose of the pictured object in every frame. In this section, we first briefly describe the method we use for tracking and segmenting the target object in videos~(Section~\ref{sec:segmentation}): From the tracking part, we obtain 2D bounding boxes of the object, which we use to define the input to our network for motion prediction. The segmentation allows us to remove the background, as we noticed it has an adverse influence on motion prediction.

\begin{figure}[t]
\setlength\tabcolsep{0pt}
\begin{center}
\begin{tabular}{c}
    \includegraphics[width=1\linewidth]{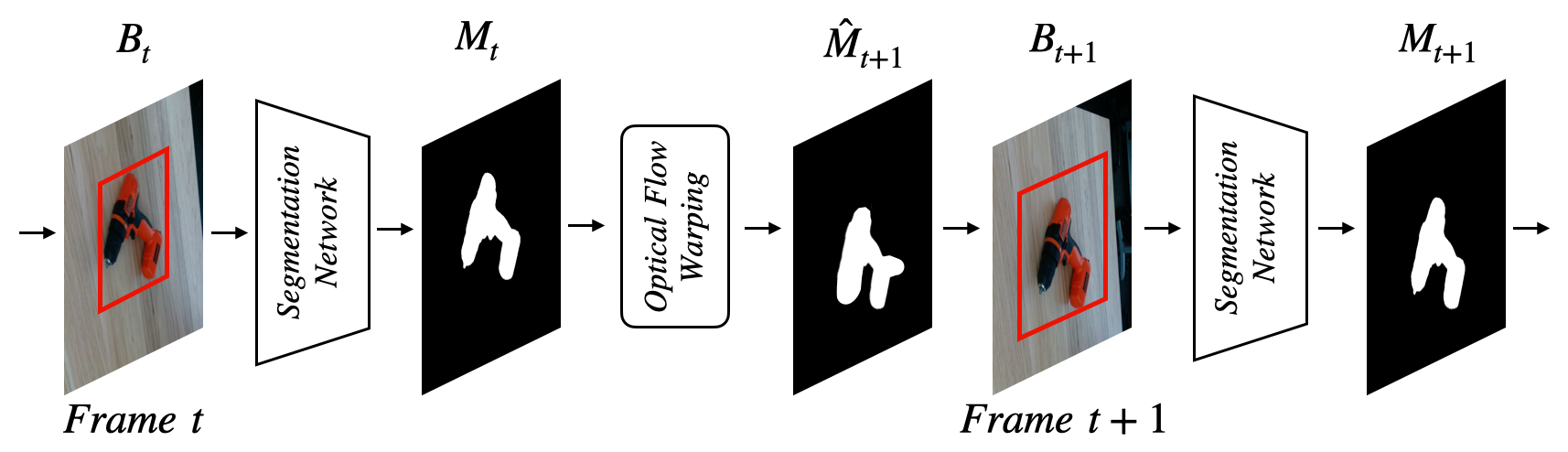}
\end{tabular}
\end{center}
\vspace{-8.5mm}
\caption{\textbf{Tracking and segmenting the target object.} 
Segmenting the object to discard background is important for good performance in complex scenes. From a 2D bounding box $B_t$ for the object at time $t$, we obtain its mask $M_t$ using a foreground-background segmentation network. We then use optical flow to predict a first estimate $\hat{M}_{t+1}$ of its mask in the next frame. From $\hat{M}_{t+1}$, we estimate its new bounding box $B_{t+1}$, on which we run the segmentation network to obtain a better mask $M_{t+1}$. This simple process is iterated over the sequence and was sufficient for our experiments.}
\label{fig:tracking}
\end{figure}
\begin{figure}[t]
\setlength\tabcolsep{0pt}
\begin{center}
\begin{tabular}{c}
    \includegraphics[height=0.23\linewidth]{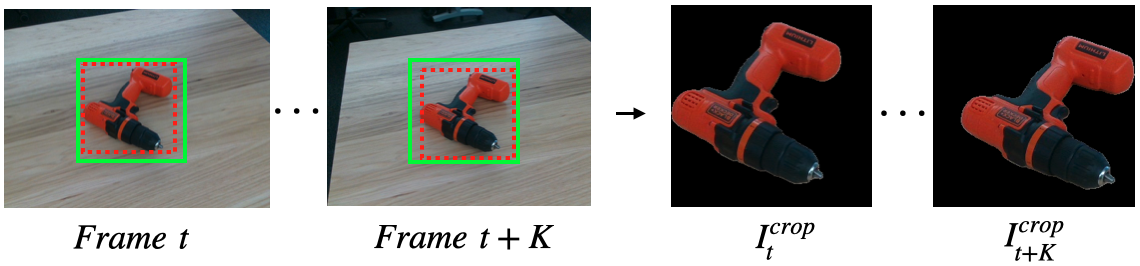}
\end{tabular}

\end{center}
\vspace{-6.5mm}
\caption{Left: We apply the same crop (in \textbf{\color{green}green}) to the $K$  consecutive input images, which keeps the smallest box containing all extracted bounding boxes from $B_t$ to $B_{t+K}$ (in \textbf{\color{red}red}).  Right: the resulting images, after cropping and background masking.}
\label{fig:crops}
\end{figure}

Then, we describe the parameterizations for the 3D translation~(Section~\ref{sec:Trans3D}) and for the 3D rotation~(Section~\ref{sec:Rot3D}) that we propose for our problem. These parameterizations are important because they are adapted to the prediction of the 6D pose in our context, and our networks rely on them to predict the 6D poses. We then detail the architecture we use for multiple frame input~(Section~\ref{sec:network}).

\subsection{2D Unseen Object Tracking and Segmentation}
\label{sec:segmentation}
We rely on a simple method for estimating 2D bounding boxes for the target object over an input video. We also obtain 2D masks for the object. A more sophisticated tracking method could be developed to prevent drift if it occurs, but it  was sufficient for our experiments, thanks to the performance of the segmentation network from~\cite{Du20211stPS}. 

This method is illustrated in Fig.~\ref{fig:tracking}. We start from a given 2D bounding box $B_0$ for the object in the first frame: This bounding box indicates to the method which object should be tracked and segmented. \cite{Du20211stPS} gives us the object mask $M_0$ in this bounding box. 

By warping the mask $M_0$ using optical flow to the second frame, we obtain a prediction $\hat{M}_1$ for the object' mask in the second frame. We use $\hat{M}_1$ to compute the bounding box $B_1$ for this frame. By applying \cite{Du20211stPS} again to $B_1$, we obtain a better mask estimate $M_1$.  We iterate this process to track the target object over the video. In practice, we use RAFT~\cite{teed2020raft} to predict the optical flow.

As shown in Fig.~\ref{fig:crops}, we feed our deep model with $K$ local frames to predict the object relative poses. These local frames are obtained by masking the background with masks $\hat{M}_1$ and applying the same cropping operation to all the frames we consider.  This cropping operation returns the smallest image region that contains all the bounding boxes $\{B_t\}_t$ for the $K$ frames.  In this way, the local frames are aligned together, which is important for the network to predict a consistent motion.

\begin{figure}
\begin{center}
\begin{tabular}{
>{\centering\arraybackslash}m{0.45\linewidth}
>{\centering\arraybackslash}m{0.45\linewidth}}
\begin{subfigure}[t]{\linewidth}
\includegraphics[width=\linewidth]{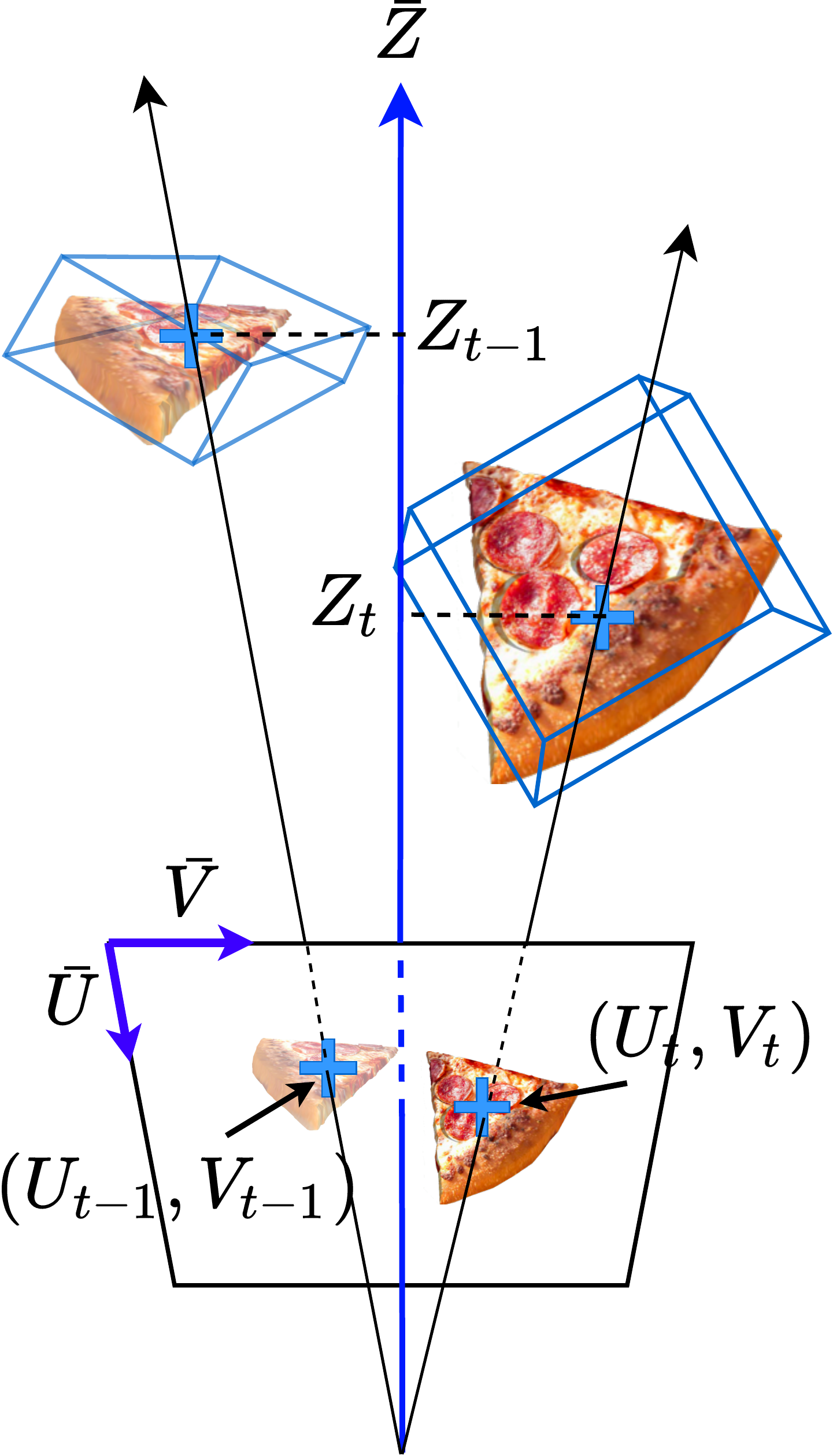}
\end{subfigure} &
\begin{subfigure}[t]{\linewidth}
\includegraphics[width=\linewidth]{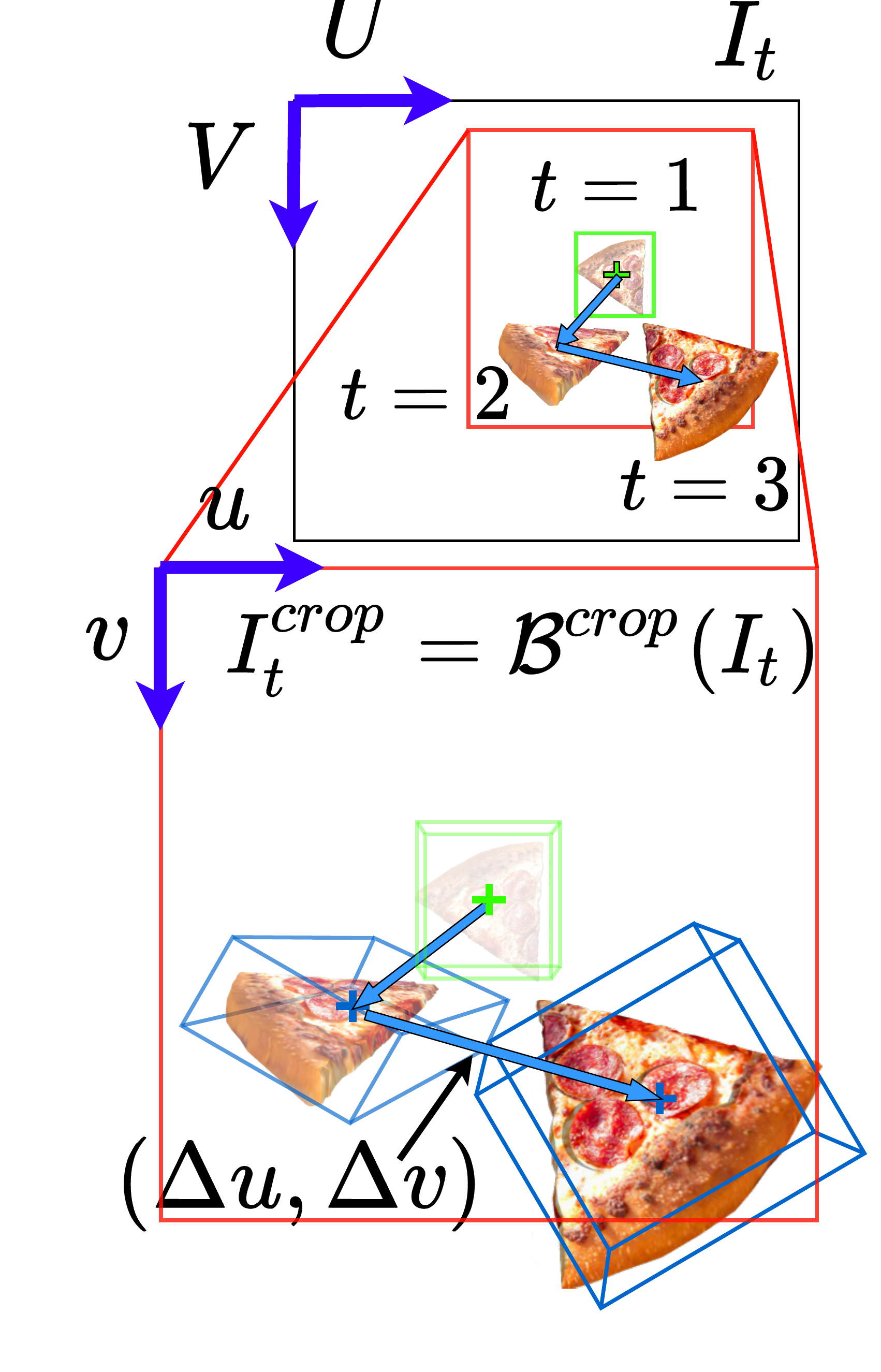}
\end{subfigure}
\end{tabular}
\end{center}
\vspace{-8mm}
\caption{\textbf{Estimating the 3D translation.} {\bf Left:} To predict the 3D object translation between two frames, we consider the 2D displacement of the projection of the object center and the scale change of its $Z$ coordinate. {\bf Right:}  The 2D displacement and the scale change are predicted from image regions extracted from the two frames image and need to be scaled.}
\label{fig:transl}
\end{figure}

\subsection{Estimating the 3D Relative Translation}
\label{sec:Trans3D}

Fig.~\ref{fig:transl} provides an overview of our parameterization of the 3D relative translation and its estimation.

\paragraph{Parameterization of 3D Relative Translation.}

The relative translation $\Delta \bT_{t-1,t} \in \IR^3$ between two frames $I_{t-1}$ and $I_t$ can be expressed as $\Delta \bT_{t-1,t} = \bT_{t} - \bT_{t-1}$, where $\bT_t$ is the 3D object center in the camera coordinate system for frame $I_t$.  In our case, we do not have access to the actual 3D size of the object, and we cannot directly estimate the norm of the relative translation $\Delta \bT_{t-1,t}$.  To introduce our translation parametrization, let's us first rewrite $\Delta \bT_{t-1,t}$ as:
\begin{equation}
\begin{aligned}
\Delta \bT_{t-1,t} &= \bT_{t} - \bT_{t-1} \\
    &= Z_t \text{K}^{-1}\begin{bmatrix} U_t \\ V_t \\ 1 \end{bmatrix} - Z_{t-1} \text{K}^{-1}\begin{bmatrix} U_{t-1} \\ V_{t-1}\\ 1 \end{bmatrix} \> ,
\end{aligned}
\label{eq:delta_T_3D}
\end{equation}
where $(U_t, V_t)$ are the pixel coordinates of the object's center reprojection and $Z_t$ its depth for frame $I_t$. $\mathrm{K}$ is the camera intrinsic matrix. During tracking on frame $I_t$, we have already estimated $(U_{t-1}, V_{t-1})$ and it is simpler to predict the 2D displacement $(\Delta U_{t-1, t}, \Delta V_{t-1, t})$ of the object center between previous frame $I_{t-1}$ and current frame $I_t$:
\begin{equation} \label{eq:delta_UV}
\begin{aligned}
    \Delta U_{t-1, t} = U_t - U_{t-1} \text{ and } \Delta V_{t-1, t} = V_t - V_{t-1} . \>
\end{aligned}
\end{equation}
Moreover, estimating the absolute depth value $Z_t$ or absolute depth variation $\Delta Z_{t-1,t}$ is also not possible in our case, we also introduce the normalized depth offset $S_{t-1,t}$ in log-scale as:
\begin{equation} \label{eq:delta_Z}
  S_{t-1,t} =  \frac{Z_t}{Z_{t-1}} - 1 \> .
\end{equation}
Note that in practice, $S_{t-1,t}$ varies around 0.

In practice, it is important to restrict the input images to a window more or less centered on the object for pose prediction: This cropping operation removes most of the background, which is irrelevant and a nuisance to pose estimation. It has however an influence on our translation parameterization.  Indeed, the cropped regions have to be scaled to use them as input to our network, which expect images of fixed resolution.  The network predicts a 2D displacement $(\Delta u, \Delta v)$ in the scaled cropped regions, and the original displacement in the image frame can be retrieved by:
\begin{equation} \label{eq:UV_uv}
    \Delta U = \alpha_u \mathrm{W}^\crop \Delta u \quad \textrm{and} \quad 
    \Delta V = \alpha_v \mathrm{H}^\crop \Delta v \> ,
\end{equation}
where $\alpha_v$ and $\alpha_v$ are the scale factors along the $u$ and $v$ axes of the images between the cropped regions and the network inputs. 
The network also predicts a depth offset $s$ for the scaled cropped regions, which also needs to be rescaled, and we use:
\begin{equation}
S = \frac{\alpha_u + \alpha_v}{2} s \> .
\end{equation}

\paragraph{Recovering $\Delta \bT_{t-1,t}$ and $\bT_t$.} If $\bT_{t-1}$ is known, we can estimate $\Delta \bT_{t-1,t}$ and $\bT_t$ from $\Delta U$, $\Delta V$, and $S$. In our experiments, and only when  comparing with previous works, we assume that the location $\bT_1$ of the object in the first frame is known in metric values, so that we can also recover the object trajectory in metric values. In our actual scenario, when no metric information is available, $Z_1$ can be set to an arbitrary positive value, and the object trajectory will be recovered up to this scale factor.

Given a frame $I_t$, together with the object center $(U_{t-1}, V_{t-1})$ in the previous frame $I_{t-1}$, we first predict $(\Delta u_{t-1, t}, \Delta v_{t-1, t}, s_{t-1, 1})$ from the cropped images $I^\crop_{t-1}$ and $I^\crop_{t}$. Then, after conversion from local coordinates to pixel coordinates using Eq.~\eqref{eq:UV_uv}, we obtain the object center $(U_t, V_t)$ from Eq.~\eqref{eq:delta_UV}, and depth value $Z_t$ from Eq.~\eqref{eq:delta_Z}. We finally retrieve the 3D translation vector between the object poses in the two frames with Eq.~\eqref{eq:delta_T_3D}.

\paragraph{Translation Loss.} The 3D translation of the target object between two frames is parameterized as a 2D displacement $(\Delta u, \Delta v)$ expressed in the cropped region frames and the depth scaling factor $s$. The training loss is implemented as:
\begin{equation} \label{eq:loss_T}
    \calL_{\Delta\bT} = \ell_1 (\Delta u, \Delta u^*) + \ell_1 (\Delta v, \Delta v^*) + \lambda \ell_1 (s, s^*) \> ,
\end{equation}
where $\Delta u^*$, $\Delta v^*$, and $s^*$ are ground truth values, $\ell_1$ is the smooth-L1 loss, and weight $\lambda$ is set to 1 in practice. 

\begin{figure}
\setlength\tabcolsep{0pt}
\begin{center}
\begin{tabular}{c}
    \includegraphics[height=0.83\linewidth]{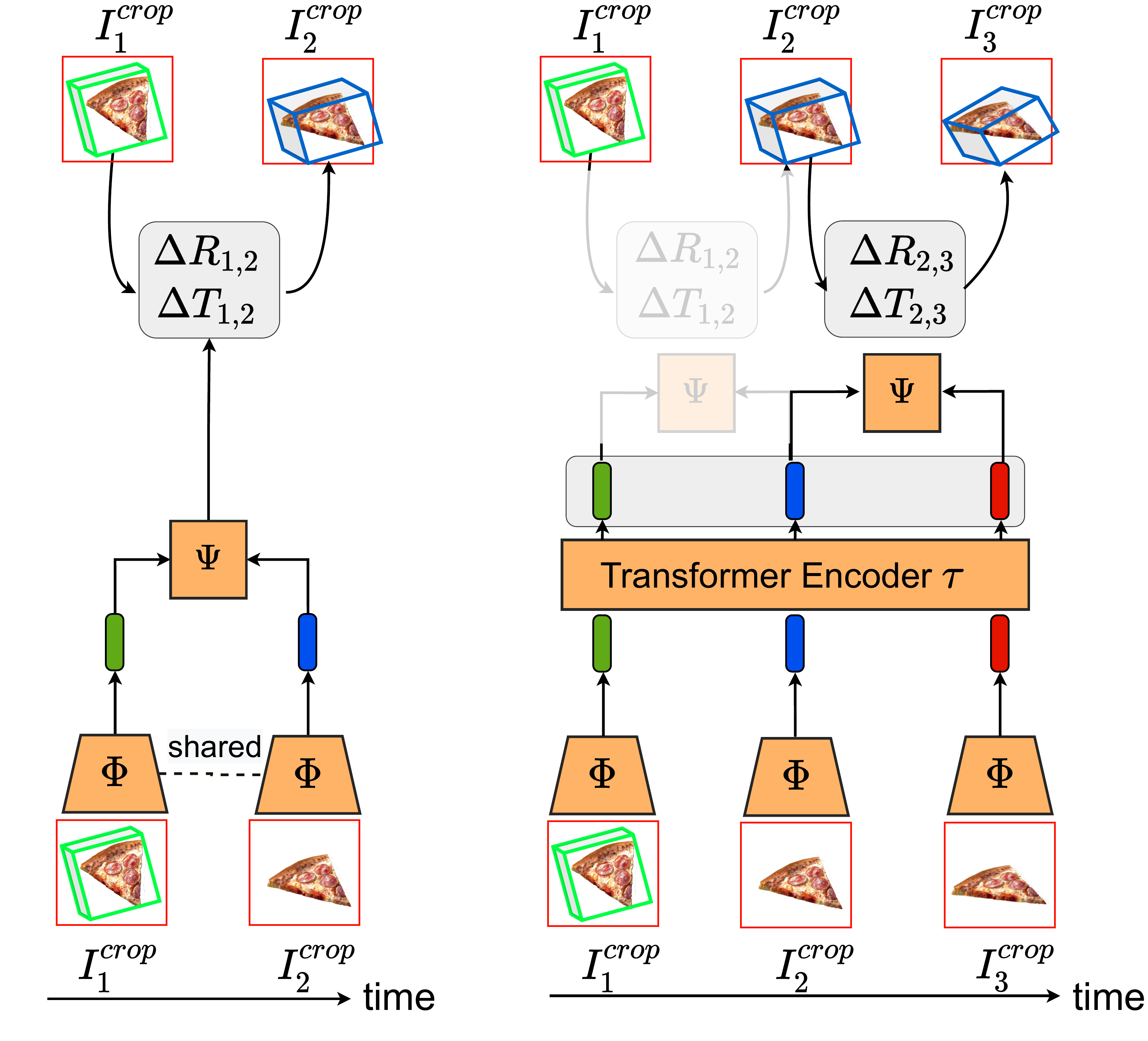}
\end{tabular}
\end{center}
\vspace{-8.5mm}
\caption{\textbf{The two architectures we consider:}
Our baseline two-frame architecture is a simple CNN-based architecture that predicts the 3D motion between two frames from these two frames only (left); Our multi-frame architecture relies on a Transformer encoder to exploit efficiently more frames as input (right). It accepts an arbitrary number of past frames as input to predict the 3D motion between the previous and current frames. All learnable components (in orange) are trained end-to-end. }
\label{fig:architecture}
\end{figure}

\subsection{Estimating the 3D Relative Rotation}
\label{sec:Rot3D}

\paragraph{Parameterization of the 3D rotation.} The relative rotation is much simpler to handle than the relative translation as it does not depend on the actual size of the object nor the scaling of the images used as input to the network. The relative rotation $\Delta \bR_{t-1, t} \in SO(3)$ between $I_{t-1}$ and $I_t$ is defined as:
\begin{equation}
    \Delta \bR_{t-1,t} = \bR_t \bR_{t-1}^\top \> ,
\label{eq:relative_rotation}
\end{equation}
where $\bR_t$ is the 3D object rotation in the camera coordinate system.
The rotation $\bR$ can be parameterized in different ways. After conducting a thorough comparative study, we opted for the axis-angle representation. This representation is defined by a vector $\bomega \in \IR^3$, where $\|\bomega\|$ and $\bomega/\|\bomega\|$ define the rotation angle and axis, respectively. More details can be found in the supplementary material.

\vspace{-3mm}
\paragraph{Rotation Loss.}
We train our networks using a simple L2 loss on the axis-angle representation:
\begin{equation} \label{eq:loss_R}
\calL_{\bomega} = \|\bomega - \bomega^*\|^2 \> ,
\end{equation}
where $\bomega^*$ denotes the ground truth axis-angle.

\vspace*{-12pt}

\paragraph{Recovering $\bR_t$.} If $\bR_{t-1}$ is known, we can recover $\bR_t$ once we estimated $\Delta \bR_{t-1,t}$. In our experiments, for comparison with other works, we will assume that the initial rotation $\bR_1$ is known for the first frame of the trajectory. This is however only for comparison purposes. In our scenario, when no 3D model is available to define a coordinate frame, the rotation for the first frame can be set to an arbitrary rotation, for example the Identity matrix. The object trajectory will be recovered up to this rotation.
\begin{algorithm}[!t]
\small
\SetKwInOut{Input}{Input}
\SetKwInOut{Init}{Initialize}
\SetAlgoLined
\SetKwProg{myproc}{procedure}{}{}
\KwResult{$\{(\bR_i, \bT_i), \textrm{for all images } I_i \textrm{ in }\mathcal{S}\}$}
\Input{
Input image sequence $\mathcal{S}=\{I_1, ..., I_N\}$\; \\ 
Object initial 2D center $(U_1, V_1)$\; \\ 
Object initial rotation $\bR_1$\; \\ 
Object initial depth $Z_1$\;\\
}

\For{$t \in [2;N[$}{

$[\feat_1, ..., \feat_n] = [\encoder(I_1), ..., \encoder(I_n)]$\;
  \If{Multiple Frames with Transformer $\transformer$}{
    $[\feat_1, ..., \feat_n] \leftarrow \transformer(\feat_1, ..., \feat_n)$ \tcp{Eq.\eqref{eq:transform_features}}
   }
   \tcp{See Sec.\ref{sec:Trans3D} and Sec.\ref{sec:Rot3D}}
 $(\bomega, \Delta u, \Delta v, s) = \Psi(\feat_{t-1}, \feat_t)$\tcp{Eq.\eqref{eq:relative_pose_network}}
 $\bT_{t-1} = Z_{t-1}\text{K}^{-1} [U_{t-1}, V_{t-1}, 1]^T$\;
 $\Delta U, \Delta V \leftarrow $\tcp{Use Eq.\eqref{eq:UV_uv}}
 $U_t, V_t \leftarrow $\tcp{Use Eq.\eqref{eq:delta_UV}}
 $Z_t \leftarrow $\tcp{Use Eq.\eqref{eq:delta_Z}}
 $\Delta \bT \leftarrow $\tcp{Use Eq.\eqref{eq:delta_T_3D}}
 $\Delta \bR \leftarrow $\tcp{Use Rodrigues Formula on $\bomega_{\Delta \bR}$}
 $\bR_{t} \leftarrow (\Delta \bR) \bR_{t-1}$\;
 $\bT_{t} \leftarrow \bT_{t-1} + \Delta \bT$\;
}
 \caption{{\bf Computing the object trajectory for a given image sequence $\mathcal{S}$ with our Transformer-based architecture.}
 The object center $(U_1, V_1)$ is provided by the user. The initial rotation can be initialized to the identity, and $Z_1$ to an arbitrary positive value. For the purpose of comparison with previous work, we initialize them to ground truth value as they do. }
 \label{algo:sequence_tracking}
\end{algorithm}
\subsection{Proposed Architecture}
\label{sec:network}

In order to efficiently incorporate information across multiple frames, and as illustrated in Fig.~\ref{fig:architecture}, our architecture takes a sequence of frames as input to predict the relative pose between the two last frames.  It first converts frames $\{ I_{t-N}, \dots, I_t \}$ into a set of image features $\{\feat_{t-i}=\encoder(I_{t-i}), i=1..N \}$ using an image embedding module $\encoder$ implemented as a ResNet18. It then projects these image features into another embedding space where interframe correlation is enhanced by the self-attention mechanism of the Transformer encoder $\transformer$:
\begin{equation}\label{eq:transform_features}
    \{ \feat_{t-N}, \dots, \feat_t \} \leftarrow \transformer \big( \feat_{t-N} \dots \feat_t \big) \> .
\end{equation}
We then predict the relative pose between $I_{t-1}$ and $I_t$ using %
\begin{equation} \label{eq:relative_pose_network}
    (\bomega_{{t-1, t}}, \Delta u_{t-1, t}, \Delta v_{t-1, t}, s_{t-1, t}) = \regressor (\feat_{t-1}, \feat_t ) \> ,
\end{equation}
where the pose regressor $\regressor(\cdot)$ is implemented as two 3-layer MLPs $\Psi_\bR$ and $\Psi_\bT$, respectively for rotation and translation estimation. We detail the architecture of these MLPs in the supplementary material. We train the different modules jointly in an end-to-end fashion.

We also consider a baseline that takes information extracted from only two input frames and which relies on a more standard deep architecture.  As our experiments will show, the Transformer-based architecture achieves significantly better performances compared to this baseline.

\paragraph{6D Tracking in sequences.} Finally, our 6D tracking algorithm, which we summarize in Algorithm~\ref{algo:sequence_tracking}, returns a sequence of 6D poses by chaining relative poses predicted from consecutive frames.

\section{Experiments}
\label{sec:experiments}
We conduct extensive experiments to evaluate the generalization capabilities of our method to unseen categories. 
We compare with CAD-based methods: DeepIM~\cite{li-eccv18-deepim}, MultiPath Learning~\cite{Sundermeyer2020MultiPathLF}, MoveIt~\cite{busam2020moveit}, and to RGB-D based methods: LatentFusion~\cite{park2020latent} and Laval~\cite{garon2018framework}. The latter uses both CAD and RGB-D images. LatentFusion requires not only the depth maps but also reference images to obtain 3D features before pose estimation. The main characteristics of all synthetic and real-world datasets we consider are summarized in Table~\ref{tab:datasets}.

We provide here qualitative examples, but much more results can also be found in the supplementary material.

\begin{table}[t]
\definecolor{Gray}{gray}{0.85}
\newcolumntype{g}{>{\columncolor{Gray}}c}
\begin{center}
\scalebox{0.70}{
\setlength\tabcolsep{5pt}
\begin{tabular}{c l r r c}
    \toprule
    Domain & Name & \#Training images & \#Test images & Videos\\
    \hline
    \multirow{2}{*}{Synthetic } & ModelNet & 420'400 & 35'000 &  \\
    & ShapeNet & 400'000 & 100'000 & \checkmark\\
    \hline
    \multirow{3}{*}{Real-world} & BOP & 22'500 & 0 & \checkmark\\
    & Laval & 0 & 14'190 & \checkmark\\
    & Moped & 0 & 3'079 & \checkmark\\
\bottomrule
\end{tabular}}
\end{center}
\vspace{-16pt}
\caption{Characteristics of the datasets considered in this paper.}
\label{tab:datasets}

\end{table}

\subsection{Experimental Setup} 
\label{sec:ExpSetup}
\begin{table*}[t]
\definecolor{Gray}{gray}{0.85}
\newcolumntype{g}{>{\columncolor{Gray}}c}
\begin{center}
\scalebox{0.65}{
\setlength\tabcolsep{5pt}
\begin{tabular}{l|cccg|cccg|cccg}
    \toprule
    & \multicolumn{4}{c|}{(5$\circ$, 5cm)} & \multicolumn{4}{c|}{ADD (0.1d)} & \multicolumn{4}{c}{Proj2D (5px)} \\ \hline
    Methods& 
    \multicolumn{1}{c}{DeepIM} & \multicolumn{1}{c}{MP} &
    \multicolumn{1}{c}{Latent} &
    \multicolumn{1}{c|}{Ours 2F} & \multicolumn{1}{c}{DeepIM} & \multicolumn{1}{c}{MP} &
    \multicolumn{1}{c}{Latent} &
    \multicolumn{1}{c|}{Ours 2F}   & \multicolumn{1}{c}{DeepIM} & \multicolumn{1}{c}{MP} & \multicolumn{1}{c}{Latent} &
    \multicolumn{1}{c}{Ours 2F} \\
    \multicolumn{1}{l|}{CAD}& 
    \checkmark & \checkmark &
      &  & \checkmark & \checkmark &
      &  & \checkmark & \checkmark &  &  \\  
    \multicolumn{1}{l|}{Depth}& 
      &   &
    \checkmark &  &   &   &
    \checkmark &  &   &  & \checkmark &  \\  
    \hline 
    bathtub & 71.6 & \textbf{85.5}
    & 85.0  & 50.0 & 88.6 & 91.5 & \textbf{92.7}& 83.6 & 73.4& 80.6& \textbf{94.9}& 52.3\\
    bookshelf & 39.2 & \textbf{81.9}& 80.2 & 68.3 & 76.4 & 85.1& \textbf{91.5}& 79.4 & 88.3 & 76.3& \textbf{91.8} & 70.0\\
    guitar& 50.4 & 69.2& \textbf{73.5}& 31.2 & 69.6 & 80.5 & \textbf{83.9} & 63.7 & 77.1 & 80.1 & \textbf{96.9}& 67.2\\
    range\_hood & 69.8 & \textbf{91.0}  & 82.9 & 60.3 & 89.6& 95.0 & \textbf{97.9} & 90.9 & 70.6 & 83.9 & \textbf{91.7}& 46.9\\
    sofa & 82.7 & \textbf{91.3}& 89.9& 77.6
    & 89.5 & 95.8 & \textbf{99.7}& 92.9 & 94.2& 86.5& \textbf{97.6}& 78.6\\
    tv\_stand& 73.6 & 85.9 & \textbf{88.6}& 83.2 & 92.1& 90.9 & \textbf{97.4}& 95.3 & 76.6& 82.5 & \textbf{96.0}& 85.8\\
    wardrobe& 62.7 & 88.7& 91.7& \textbf{92.0} & 79.4 & 92.1& 97.0 & \textbf{98.0} & 70.0& 81.1& \textbf{94.2} & 85.8\\ 
    \hline
    Mean & 64.3 & 84.8&\textbf{85.5} & 66.1 & 83.6  & 90.1 & \textbf{94.3} & 87.5 & 73.3& 81.6& \textbf{94.7}& 69.5\\ 
\bottomrule
\end{tabular}}
\vspace{-16pt}
\end{center}
\caption{Comparison to DeepIM~\cite{li-eccv18-deepim}, Multi-Path Learning (MP)~\cite{Sundermeyer2020MultiPathLF} and LatentFusion (Latent)~\cite{park2020latent} on unseen categories of the ModelNet dataset. Note that DeepIM and MP require the CAD models for refinement; LatentFusion uses RGB-D images and reference frames. In all cases only two consecutive frames are used, which is why we only report results for our two-frame method.}
\label{tab:modelnet}
\end{table*}

\subsubsection{Synthetic Datasets}
We follow previous works~\cite{park2020latent} to generate a synthetic dataset from CAD models of ModelNet~\cite{Wu_2015_CVPR}. The target view is rendered at constant translation $t = (0,0,500~mm)$ and random rotation $R$. Then, we render another  view with that pose perturbed by angles $\beta_{\{x|y|z\}} \sim \mathcal{N}(0,\,(15^o)^2)$ for each rotation axis and a translational offset $(\Delta x, \Delta y, \Delta z)$ with $\Delta_{\{x|y\}} \sim \mathcal{N}(0, 10^2)$, $\Delta z \sim \mathcal{N}(0, 50^2)$ (mm) with the total angular perturbation lower than $45^o$.

Previous methods only evaluate their models with pairs of images, we generalize this setting to multi-frames with CAD models from ShapeNet to evaluate the performance of the methods on long sequences. To do so, we create new synthetic videos by following \cite{garon2018framework}: the first frame of the video is rendered with a random translation $t = (0,0,Z)$ (mm) with Z sampled uniformly from 0.4m to 2m and a random rotation $R \sim SO(3)$. The next frame is rendered by perturbating the object pose in the previous frame by angles $\beta_{\{x|y|z\}} \sim \mathcal{N}(0,\,(20^o)^2)$ and a translation$\Delta_{\{x|y|z\}} \sim \mathcal{N}(0, 20^2)$ (mm). We iterate the process until we get 100 frames for each CAD model. More detailed about synthetic data generation can be found in the supplementary material.

\vspace{-3mm}
\subsubsection{Real-world Datasets}
In addition to the synthetic datasets, we evaluate our methods on two different real-world datasets: 1) the Laval dataset~\cite{garon2018framework}, which contains 297 sequences of 11 real objects captured in 3 different scenarios. As the {\it interaction} scenario of Laval~\cite{garon2018framework} requires hand occlusion which is a very specific setting, we follow  MoveIt~\cite{busam2020moveit} by focusing more on the {\it occlusion} scenario. Note that MoveIt only evaluates on 8 of the 11 objects of Laval, while we evaluate on all 11 objects and report the averaged performance; 2) the MOPED dataset~\cite{park2020latent}, which consists of 11 household objects covering all views of the objects.

\vspace{-3mm}
\subsubsection{Implementation Details}
We implemented our models using the Pytorch~\cite{paszke2017automatic} framework.
The image encoding module $\encoder$ is a ResNet18~\cite{He2016ResNet} which extracts a global embedding of dimension 256 for each input image resized to $224 \times 224$. The pose regressor $\regressor$ is composed of two identical MLPs with three hidden layers of size 800-400-200. The Transformer encoder $\transformer$ is implemented with a single multi-head self-attention layer with 12 heads, and 512 hidden neurons in the feedforward~(MLP) layer.
More details are given in the supplementary material.

\begin{table*}[t]
\definecolor{Gray}{gray}{0.85}
\begin{center}
\scalebox{0.65}{
\setlength\tabcolsep{5pt}
\begin{tabular}{cc|cccccccc}
    \toprule
    &  & Bathtub & Bookshelf & Bus & Car & Cabinet & Display & Telephone & Mean \\ 
    \hline
    \multirow{3}{*}{\bf Translation (mm)} & Avg signed motion & 24.1 & 24.1 & 24.5 & 23.8 & 24.7 & 23.9 & 25.3 & 24.3 \\
    & \cellcolor[gray]{0.85} Ours 2F & \cellcolor[gray]{0.85} 14.1 & \cellcolor[gray]{0.85} 14.9 & \cellcolor[gray]{0.85} 14.3 & \cellcolor[gray]{0.85} 13.6 & \cellcolor[gray]{0.85} 13.4 & \cellcolor[gray]{0.85} 15.27 & \cellcolor[gray]{0.85} 15.3 & \cellcolor[gray]{0.85} 16.2 \\
    & \cellcolor[gray]{0.85} Ours MF &  \cellcolor[gray]{0.85} \bf 13.1 & \cellcolor[gray]{0.85} \bf 13.8 & \cellcolor[gray]{0.85} \bf 13.1 & \cellcolor[gray]{0.85} \bf 11.8 & \cellcolor[gray]{0.85} \bf 12.8 & \cellcolor[gray]{0.85} \bf 14.2 & \cellcolor[gray]{0.85} \bf 13.6 & \cellcolor[gray]{0.85} \bf 13.2 \\  
    \hline
    \multirow{3}{*}{\bf Rotation (deg)} & Avg signed motion & 20.8 & 20.5 & 20.8 & 20.7 & 20.6 & 21.0 & 21.0 & 20.7 \\
    & \cellcolor[gray]{0.85} Ours 2F & \cellcolor[gray]{0.85} 5.4 & \cellcolor[gray]{0.85} 5.2 & \cellcolor[gray]{0.85} 5.4 & \cellcolor[gray]{0.85} 3.8 & \cellcolor[gray]{0.85} 5.6 & \cellcolor[gray]{0.85} 6.3 & \cellcolor[gray]{0.85} 5.7 & \cellcolor[gray]{0.85} 5.3 \\
    & \cellcolor[gray]{0.85} Ours MF & \cellcolor[gray]{0.85} \bf 5.4 & \cellcolor[gray]{0.85} \bf 4.8 & \cellcolor[gray]{0.85} \bf 4.8 & \cellcolor[gray]{0.85} \bf 3.3 & \cellcolor[gray]{0.85} \bf 4.5 & \cellcolor[gray]{0.85} \bf 5.9 & \cellcolor[gray]{0.85} \bf 5.6 & \cellcolor[gray]{0.85} \bf 4.9 \\
\bottomrule
\end{tabular}
}
\end{center}
\vspace{-18pt}
\caption{Comparaison of our two-frame (Ours 2F) and our multi-frames (Ours MF) architectures on synthetic videos of ShapeNet. ``Average signed motion'' gives the average motion over all the sequences of 15 frames in our synthetic videos. Ours MF outperforms Ours 2F on all the categories.}
\label{tab:shapenet}
\end{table*}

\vspace{-5mm}
\subsubsection{Evaluation Metrics}
We use five evaluation metrics as done in \cite{li-eccv18-deepim,Sundermeyer2020MultiPathLF,park2020latent,busam2020moveit,garon2018framework}: 1) $(k^\circ, k~\text{cm})$: a pose is considered correct if the geodesic error and translation error are within $k^\circ$ and $k$ cm. 2) ADD~\cite{Hinterstoier2012ModelBT}: the average distance between points after being transformed by the ground truth and predicted poses. 3) ADD-S: a modification to ADD that computes the average distance to the closest point rather than the ground truth point to account for symmetric objects. 4) Proj.2D: the pixel distance between the projected points of the ground truth and predicted pose. 5) the geodesic error and translation error between the prediction and the ground-truth for every sequence of 15 frames as done in \cite{garon2018framework,busam2020moveit}.
\begin{figure}[!t]
\newlength{\qualmodelnetheight}
\setlength\qualmodelnetheight{1.6cm}
\setlength\lineskip{1pt}
\setlength\tabcolsep{1pt}
\begin{center}
\begin{tabular}{ccccc}
 \includegraphics[ height=\qualmodelnetheight, width=\qualmodelnetheight]{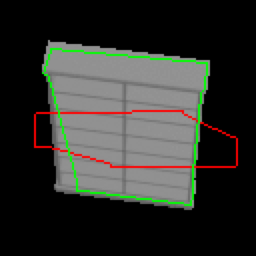} &
 \includegraphics[ height=\qualmodelnetheight, width=\qualmodelnetheight]{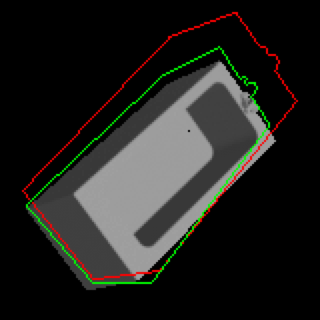} &
 \includegraphics[ height=\qualmodelnetheight, width=\qualmodelnetheight]{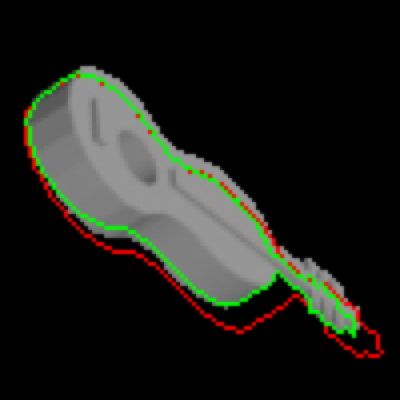} &
 \includegraphics[ height=\qualmodelnetheight, width=\qualmodelnetheight]{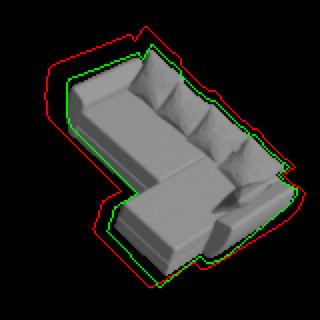}&
 \includegraphics[ height=\qualmodelnetheight, width=\qualmodelnetheight]{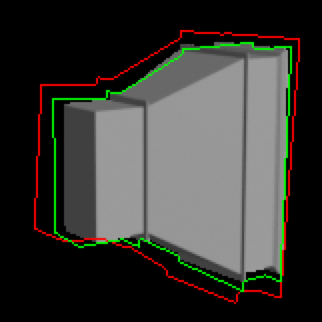} 
\end{tabular}\\
\begin{tabular}{c}
\includegraphics[ height=0.2\teaserheight]{images/introduction/time_arrow.pdf}
\end{tabular}
\end{center}
\vspace{-8mm}
\caption{\textbf{Qualitative results on ModelNet.} We show the target image with an initial perturbed pose in a \textbf{\color{red}red} wireframe and our pose prediction with a \textbf{\color{green}green} wireframes. The object wireframes are used only for visualization. }
\label{fig:modelnet_qualitative_results}
\end{figure}

\newcommand{\numbf}[1]{\bf #1}
\begin{table}[h]
\definecolor{Gray}{gray}{0.85}
\begin{center}
\scalebox{0.61}{
\setlength\tabcolsep{5pt}
\begin{tabular}{lcc|cccc|cccc}
\toprule
    \multirow{3}{*}{\bf Method} & \multirow{3}{*}{\bf CAD} &
    \multirow{3}{*}{\bf Depth} &
    \multicolumn{4}{c|}{\textbf{Translation (mm)}}  & \multicolumn{4}{c}{\textbf{Rotation (deg)}}\\
     & & & \multicolumn{4}{c|}{Occlusion} & \multicolumn{4}{c}{Occlusion} \\
    & & & $0\%$ & $15\%$ & $30\%$ & $45\%$ & $0\%$ & $15\%$ & $30\%$ & $45\%$ \\
     \hline
    
    Laval ~\cite{garon2018framework} &\checkmark & \checkmark & \numbf{6.7} & 11.1 & 18.9 & 25.9 & \numbf{5.3} &  \numbf{8.4}& \numbf{16.1} & 26.8\\
             
    MoveIt~\cite{busam2020moveit}  & \checkmark &   & 14.5 & 13.9 & 25.1 & 22.5
             & 33.3 & 31.8 & 34.1 & 28.86\\

    \rowcolor{Gray} Ours 2F & &
             & 16.18 & 14.92 & 15.38  & 15.44
             & 19.83 & 20.78 & 20.16 & 20.33\\
    \rowcolor{Gray} Ours MF & &
             & 9.82 & \numbf{10.27} & \numbf{9.97} & \numbf{10.16}
             & 16.41 & 16.13 & 16.21 & \numbf{17.76}\\ 
\bottomrule
\end{tabular}}
\end{center}
\vspace{-12pt}
\caption{Comparison with 3D tracking methods Laval generic~\cite{garon2018framework} and MoveIt~\cite{busam2020moveit} on the Laval dataset. Laval generic uses the object's 3D model and depth maps; MoveIt uses the 3D model. We use neither of these extra inputs. Our method is more robust to occlusion compared to other methods and achieves better superior performance on videos with severe occlusion.}
\label{tab:laval-pose}
\end{table}
\subsection{Evaluation on Synthetic Datasets}

\subsubsection{Results on ModelNet}
We compare the performance of our method with DeepIM~\cite{li-eccv18-deepim}, MultiPath Learning~(MP)~\cite{Sundermeyer2020MultiPathLF} and LatentFusion~\cite{park2020latent} on the unseen classes of ModelNet~\cite{Wu_2015_CVPR}. Both DeepIM and MP require the object's CAD model for prediction while LatentFusion relies on depth images during testing. Our method takes only RGB images as input during testing. As shown in Table~\ref{tab:modelnet}, our method can achieve comparable performance for the ADD metric while requiring much less information. Our method even achieves better results than DeepIM for the $(5^\circ, 5cm)$ and ADD metrics. Furthermore, our method estimates directly the relative pose without any iterative refinement adopted in DeepIM and MP, which demonstrates the effectiveness of our method on unseen classes. We show the output of our method on the ModelNet dataset in Fig.~\ref{fig:modelnet_qualitative_results}. 

\subsubsection{Results on ShapeNet}
We compare the performance of our two-frame architecture network which takes two frames as input with our multi-frame Transformer based network in the ShapeNet setting. As shown in Table~\ref{tab:shapenet}, our multi-frame Transformer outperforms the two-frame architecture network on every unseen class, which validates the benefit of taking multiple frames as input and demonstrates the effectiveness of our multi-frame Transformer network.

\subsection{Evaluation on Real-World Datasets}
\subsubsection{Results on Laval}
\begin{figure*}
\newlength{\quallavalheight}
\setlength\quallavalheight{1.45cm}
\newlength{\qualmopoedheight}
\setlength\qualmopoedheight{1.55cm}
\setlength\lineskip{3pt}
\setlength\tabcolsep{3pt}
\begin{center}
\begin{tabular}{cccccc}
 \includegraphics[ height=\quallavalheight]{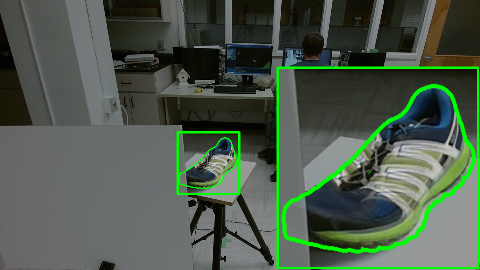} &
 \includegraphics[ height=\quallavalheight, width=\quallavalheight]{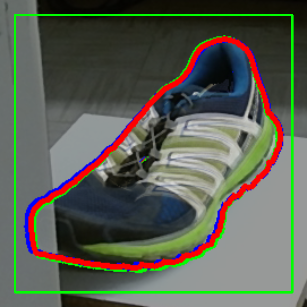} &
 \includegraphics[ height=\quallavalheight, width=\quallavalheight]{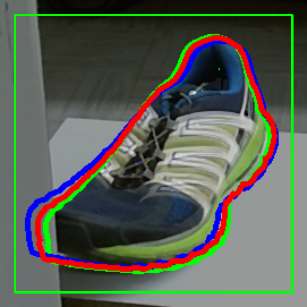} &
 \includegraphics[ height=\quallavalheight, width=\quallavalheight]{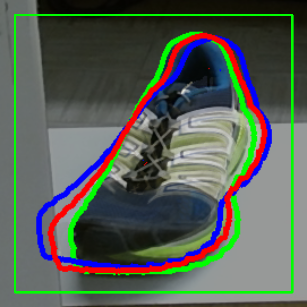}&
 \includegraphics[ height=\quallavalheight, width=\quallavalheight]{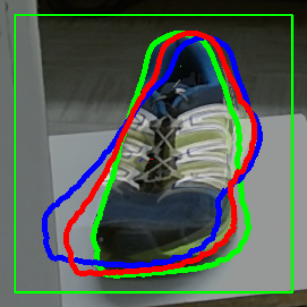}&
 \includegraphics[ height=\quallavalheight, width=\quallavalheight]{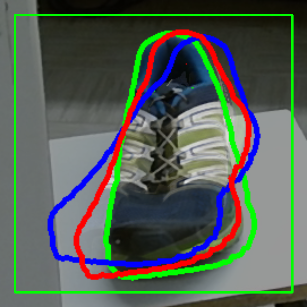}
\end{tabular}\\
\begin{tabular}{ccccc}
\includegraphics[ height=\qualmopoedheight, width=1.3\qualmopoedheight]{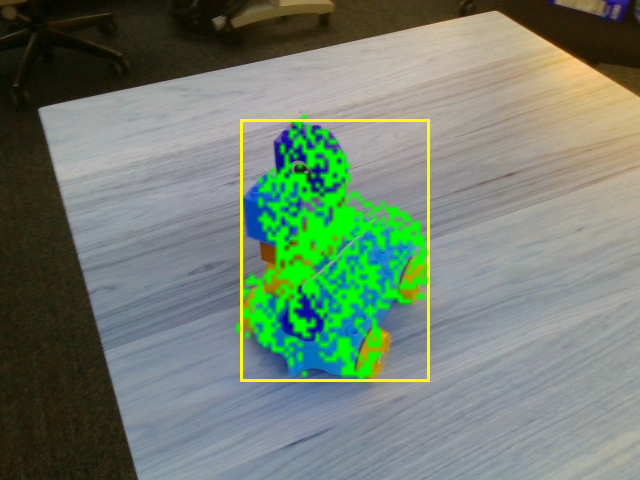} &
 \includegraphics[ height=\qualmopoedheight, width=1.3\qualmopoedheight]{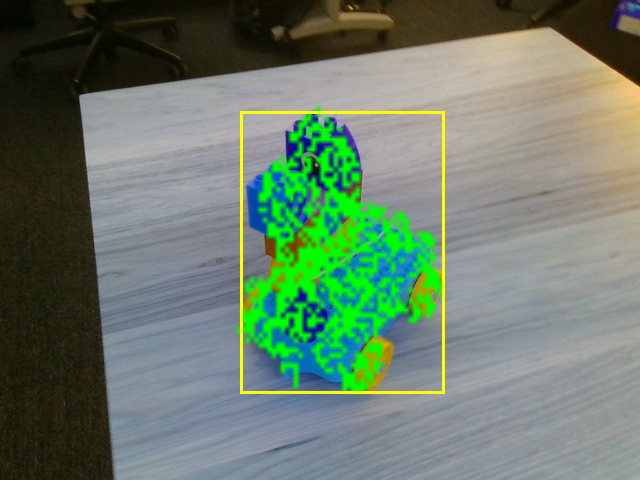} &
 \includegraphics[ height=\qualmopoedheight, width=1.3\qualmopoedheight]{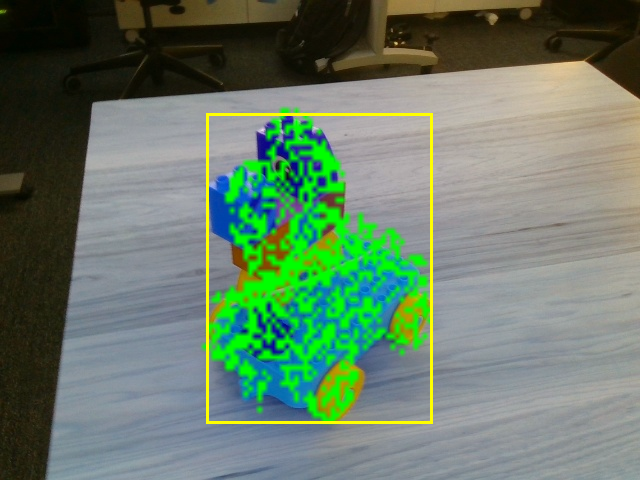}&
 \includegraphics[ height=\qualmopoedheight, width=1.3\qualmopoedheight]{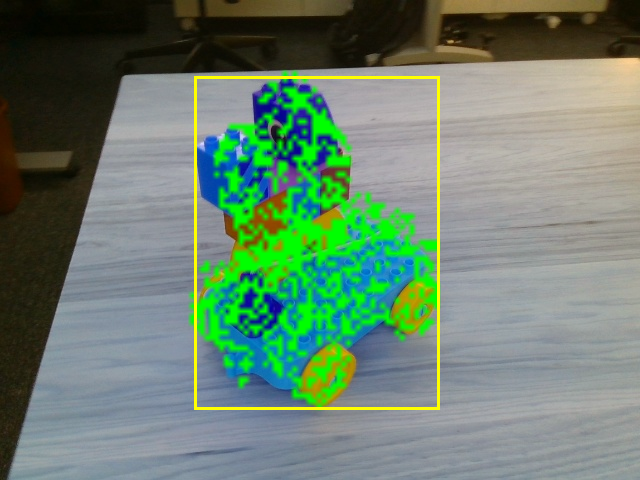}&
 \includegraphics[ height=\qualmopoedheight, width=1.3\qualmopoedheight]{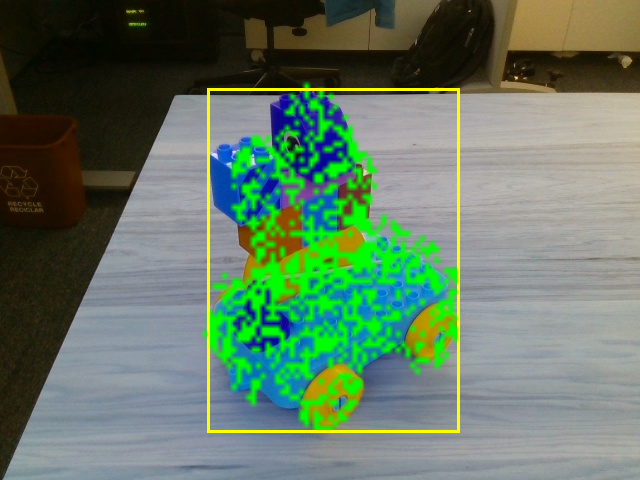}
\end{tabular}\\
\begin{tabular}{ccccc}
 \includegraphics[ height=\qualmopoedheight, width=1.3\qualmopoedheight]{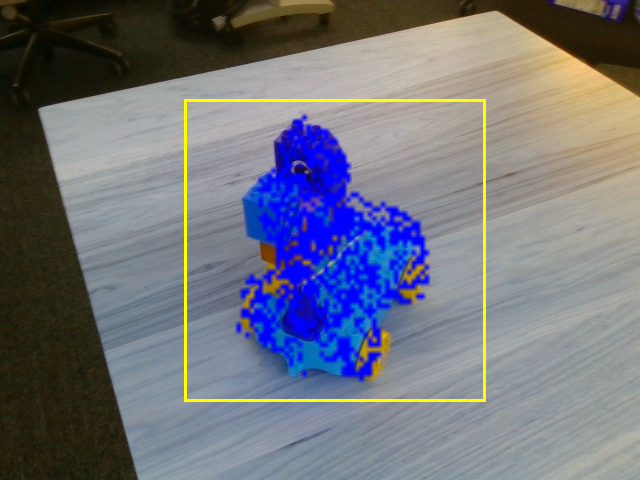} &
 \includegraphics[ height=\qualmopoedheight, width=1.3\qualmopoedheight]{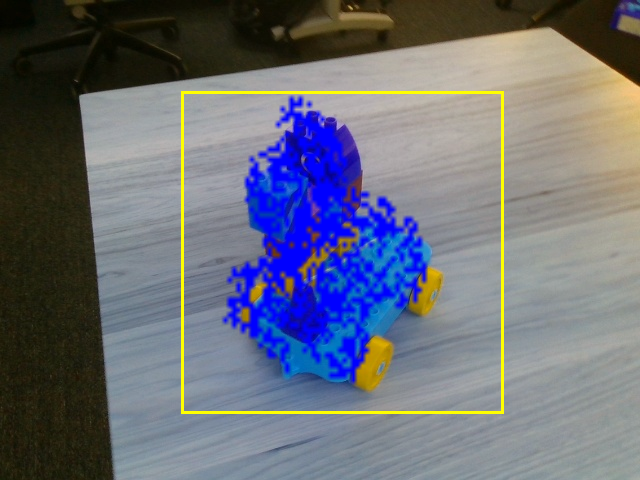} &
 \includegraphics[ height=\qualmopoedheight, width=1.3\qualmopoedheight]{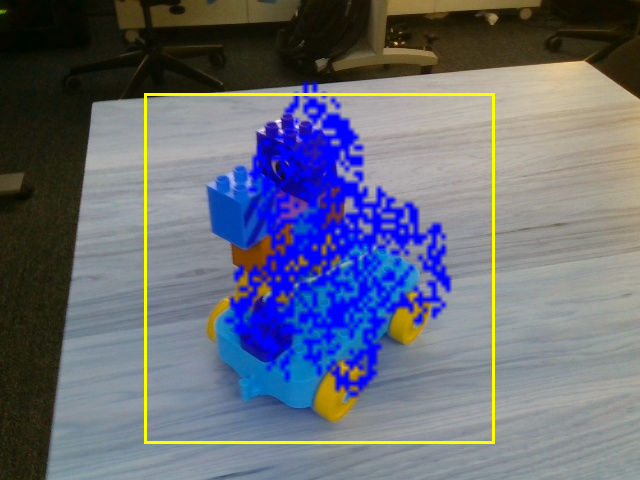}&
 \includegraphics[ height=\qualmopoedheight, width=1.3\qualmopoedheight]{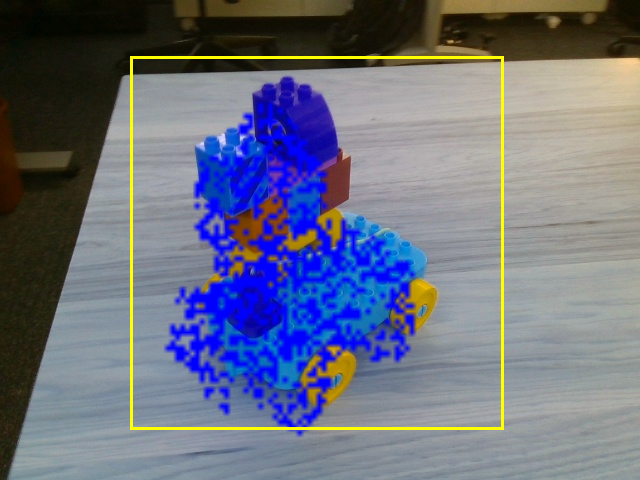}&
 \includegraphics[ height=\qualmopoedheight, width=1.3\qualmopoedheight]{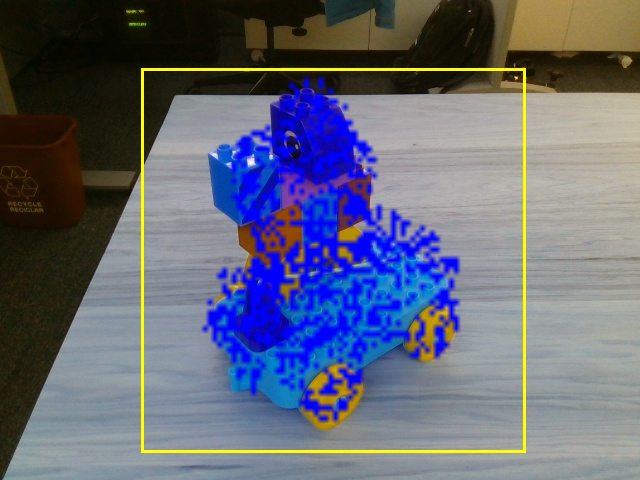}
\end{tabular}\\
\begin{tabular}{ccccc}
\includegraphics[ height=\qualmopoedheight, width=1.3\qualmopoedheight]{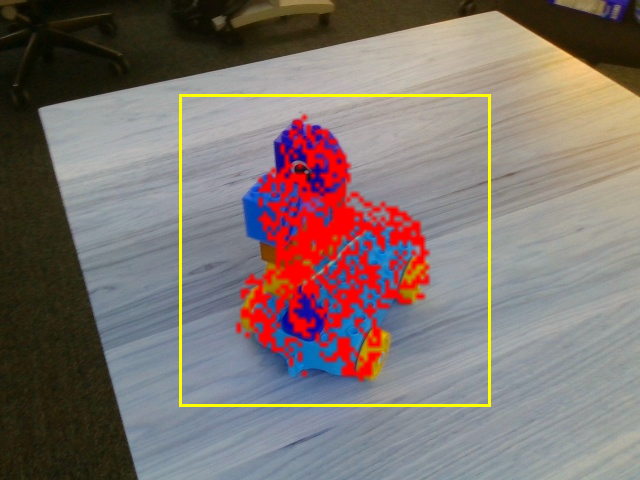} &
 \includegraphics[ height=\qualmopoedheight, width=1.3\qualmopoedheight]{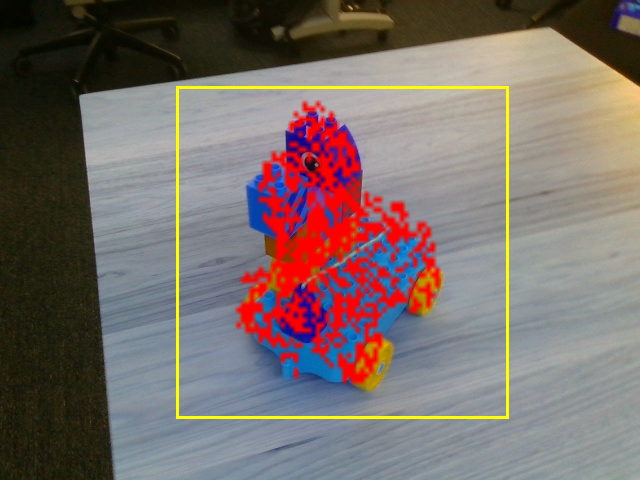} &
 \includegraphics[ height=\qualmopoedheight, width=1.3\qualmopoedheight]{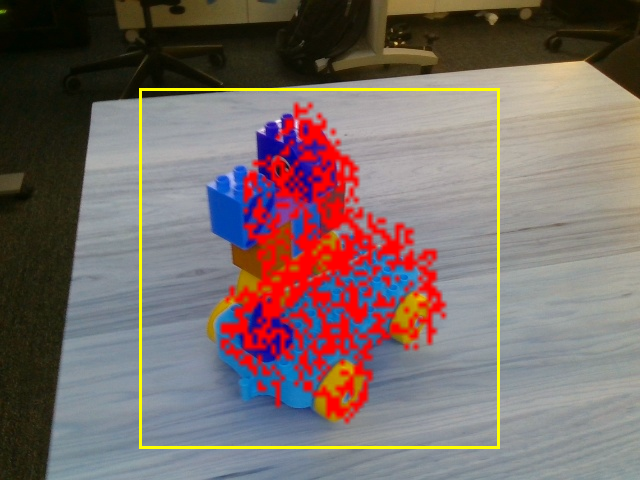}&
 \includegraphics[ height=\qualmopoedheight, width=1.3\qualmopoedheight]{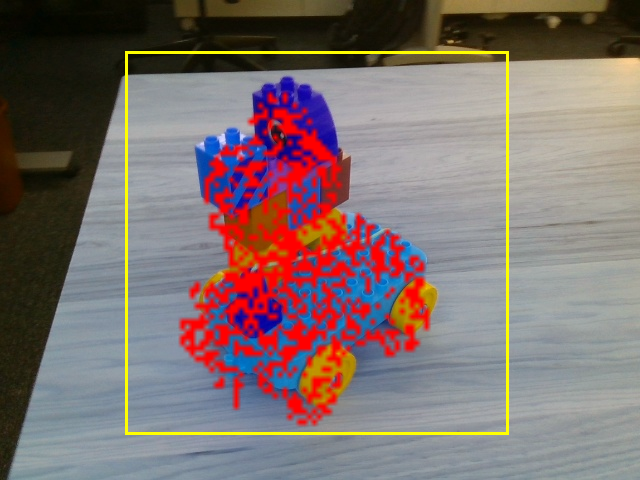}&
 \includegraphics[ height=\qualmopoedheight, width=1.3\qualmopoedheight]{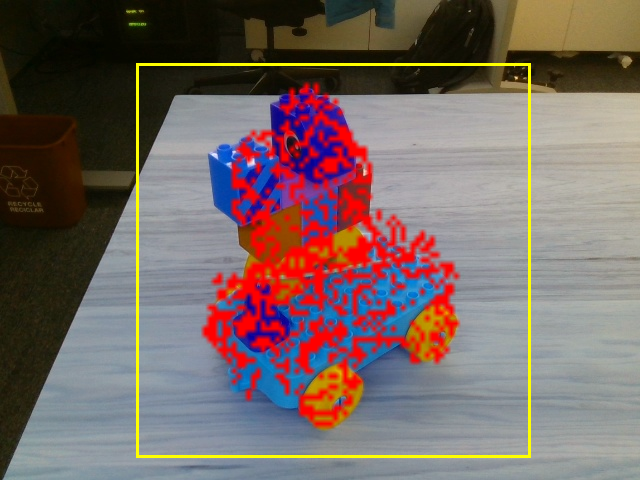}
\end{tabular}\\
\begin{tabular}{@{\;\;\;\;\;\;\;\;\;\;\;\;\;\;\;\;\;}c}
\includegraphics[height=0.25\qualmopoedheight]{images/introduction/time_arrow.pdf}
\end{tabular}
\end{center}
\vspace{-22pt}
\caption{\textbf{Qualitative results on real-world datasets.} We show in each image the ground-truth in \textbf{\color{green}green}, the prediction with Ours two-frame in \textbf{\color{blue}blue} and the prediction with Ours multi-frame in \textbf{\color{red}red}. First row shows our results on a sequence images of object “shoe" of the Laval dataset~\cite{garon2018framework}. The three last rows show our results on a sequence of images on Object “duplo\_dude" of the Moped dataset~\cite{park2020latent}.}
\label{fig:qualitative_results}

\end{figure*}

We compare our method with two recent methods: Laval-generic~\cite{garon2018framework} and MoveIt~\cite{busam2020moveit}. Following their evaluation protocol, the ground truth pose is given every 15 frames. Table~\ref{tab:laval-pose} reports the results. 
Our two-frame architecture network is comparable to other methods without using the CAD models or depth images. Our multi-frame Transformer network further boosts the performance. Our method achieves superior results especially on sequences with severe occlusion, and our method degrades very little compared with other methods when increasing the occlusion rate. Thus, our method is more robust to occlusion.
Our two models, especially the multi-frame version, achieve very good results and even often outperform the two other methods despite a clear disadvantage. The multi-frame version achieves the best overall performance on 3D relative translation, and much better results for 3D relative rotation estimation than MoveIt that uses CAD models. Fig.~\ref{fig:qualitative_results} provides some qualitative results.

\subsubsection{Results on Moped}
We also compare our methods with LatentFusion~\cite{park2020latent}, which relies on RGB-D images, on the Moped dataset. The performance of our method is comparable with LatentFusion when LatentFusion uses 1 reference image. It still has a 20\% gap with LatentFusion, when LatentFusion uses 8 reference images. We show some outputs of our method on the Moped dataset in Fig.~\ref{fig:qualitative_results}.

\subsection{Limitations}
\begin{table}
\definecolor{Gray}{gray}{0.85}
\newcolumntype{g}{>{\columncolor{Gray}}c}
\begin{center}
\scalebox{0.61}{
\setlength\tabcolsep{5pt}
\begin{tabular}{l|ccgg|ccgg}
    \toprule
    & \multicolumn{4}{c|}{AUC (ADD)$\uparrow$} & \multicolumn{4}{c}{AUC (ADD-S)$\uparrow$} \\ 
    \midrule
    Methods & 
    \multicolumn{1}{c}{Latent} & \multicolumn{1}{c}{Latent} &
    \multicolumn{1}{c}{Ours 2F} & \multicolumn{1}{c|}{Ours MF} & \multicolumn{1}{c}{Latent} & \multicolumn{1}{c}{Latent} &
    \multicolumn{1}{c}{Ours 2F} & \multicolumn{1}{c}{Ours MF} \\
    Depth & 
     \checkmark & \checkmark &  &  & \checkmark & \checkmark  &  &   \\  
    \# R. views & 1 & 8 & 1 & 1 & 1 & 8 & 1 & 1   \\   
    \hline 
    black\_drill   & - & 89.4 & 44.6 & 45.6 & - & 95.1 & 62.4 & 68.9 \\ 
    cheezit        & - & 24.6 & 44.9 & 42.0 & - & 92.2 & 72.4 & 74.4 \\
    duplo\_dude    & - & 88.9 & 38.4 & 39.1 & - & 94.7 & 62.7 & 69.7 \\
    duster         & - & 47.2 & 55.0 & 57.2 & - & 87.9 & 78.5 & 84.3 \\
    graphic\_card & - & 73.0 & 45.5 & 47.9 & - & 79.2 & 70.3 & 77.7 \\
    orange\_drill  & - & 78.3 & 49.1 & 52.2 & - & 93.5 & 64.7 & 72.7 \\ 
    pouch          & - & 54.8 & 26.7 & 23.9 & - & 91.7 & 41.5 & 42.8 \\ 
    remote         & - & 60.0 & 64.1 & 64.4 & - & 91.9 & 73.9 & 79.8 \\
    rinse\_aid     & - & 67.4 & 70.2 & 72.0 & - & 93.3 & 83.2 & 90.7 \\
    toy\_plane     & - & 82.9 & 21.4 & 22.3 & - & 92.2 & 43.9 & 50.1 \\
    vim\_mug       & - & 40.0 & 61.9 & 60.4 & - & 91.7 & 79.0 & 83.5 \\
    \hline
    \hline
    Mean  & 23.9 & 68.3 & 47.5 & 47.9 & 78.7 & 90.9 & 66.6 & 72.2 \\
\bottomrule
\end{tabular}}
\end{center}
\vspace{-3mm}
\caption{Comparison to LatentFusion~(Latent)~\cite{park2020latent} on the Moped dataset. We report the Area Under Curve~(AUC) for each metric as done in \cite{park2020latent}. Note that our method does not require any reference image. Our method is comparable with LatentFusion when LatentFusion uses one reference image.}

\vspace{-3mm}
\end{table}

\noindent Our method comes with some limitations:
\begin{itemize}[left=0pt]
    \item Estimation of $\Delta T_z$: We show in the supplementary material that the translation error along the $Z$ axis is much larger than the errors along the other two axes. This is a common problem in 6D pose estimation: Estimating motion along the $Z$ axis is very challenging for RGB-based methods, as relatively large $\Delta T_z$ can occur even with minimal appearance change between two frames.
    \item Our method may fail when our segmentation model fails to segment the target object. However, our experience is that this model is surprisingly robust, and that our pose estimation model is robust to incorrect segments.
\end{itemize}

\section{Conclusion}
\label{sec:conclusion}
In this paper, we explored what can be done for 3D object tracking in the absence of 3D model, depth data, registered images,  and even knowledge about the object class. The accuracy we  obtain is surprisingly good, as it is on par with previous methods that have access to much more information. Beyond downstream possible applications such as robotics and Man-Machine-Interaction, we hope our approach will encourage more research on the ``Zero-shot Zero-CAD'' topic.

\vspace{0.2cm}
{\small \noindent\textbf{Acknowledgments.} We thank Xi Shen and Mathis Petrovich for helpful discussions. This research was produced within the framework of Energy4Climate Interdisciplinary Center~(E4C) of IP Paris and Ecole des Ponts ParisTech, and was supported by the 3rd \emph{Programme d’Investissements d’Avenir} [ANR-18-EUR-0006-02], the support of the Chair “Challenging Technology for Responsible Energy" led by l’X – Ecole polytechnique and the Fondation de l’Ecole polytechnique, sponsored by TOTAL, and the ChistEra IPALM project. This work was performed using HPC resources from GENCI–IDRIS 2021-AD011012294.}

{\small
\bibliographystyle{ieee_fullname}
\bibliography{bibliography}
}

\clearpage
\section*{}
\textbf{\Large{Supplementary Material}}
\vspace{-2mm}

\section{Rotation representation}
\label{sec:rotation_representation}
\begin{figure}[h]
\begin{center}
	\includegraphics[width=0.97\linewidth]{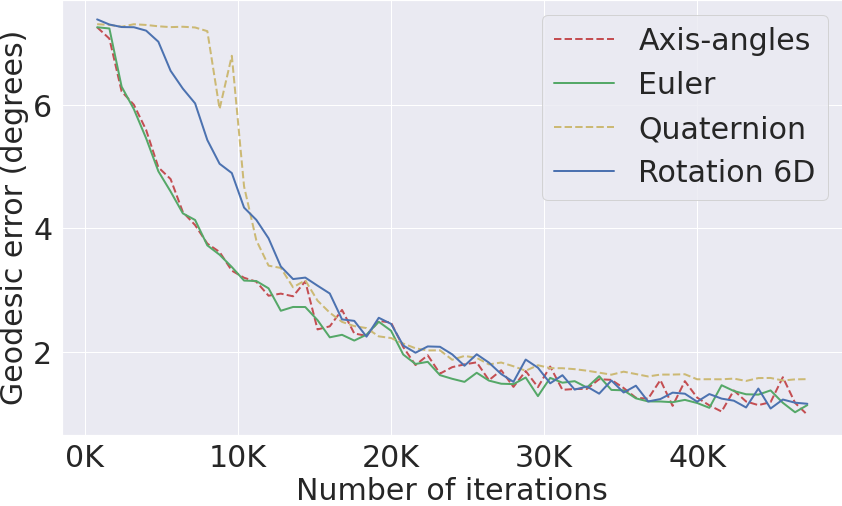}
    \vspace{-2mm}
	\caption{{\bf Ablation study on synthetic images created from CAD models of ShapeNet~\cite{chang2015shapenet} with Our two-frames.} The average of the rotational motion equal to 7.25 degrees. We show that the performance is almost the same when using different types of representation for the rotation.}
	\label{fig:rotation_representation}
\end{center}
\end{figure}

Figure~\ref{fig:rotation_representation} shows our ablation study on different representations of rotation on synthetic images created from CAD models of ShapeNet~\cite{chang2015shapenet}: Euler angles, axis-angles, a quaternion and the rotation 6D method from \cite{Zhou2019Rotation6D}. While the rotation 6D method  has been shown recently to be better than the other representations in absolute pose estimation~\cite{labbe2020}, our ablation study shows that we do not gain the performance when the motion is small such as in objet tracking.

\vspace{-3mm}
\section{Architecture details}
\paragraph{Transformer.} Our Transformer $\transformer$ is implemented with the TransformerEncoder layer of Pytorch~\cite{paszke2017automatic} with only one encoder layer with 12 heads and an MLP containing 512 hidden neurons. We do not use dropout in the encoder layer.
\vspace{-2mm}
\paragraph{Pose Regressor.} Our pose regression module $\regressor(\cdot)$ is composed of two MLPs, one for translation and one for rotation. Each MLP is made of three layers with 800, 400, 200 hidden neurons respectively. The middle layers are followed by a batch normalization and a ReLU activation.
\begin{figure}[!ht]
\begin{center}
	\includegraphics[width=8.5cm]{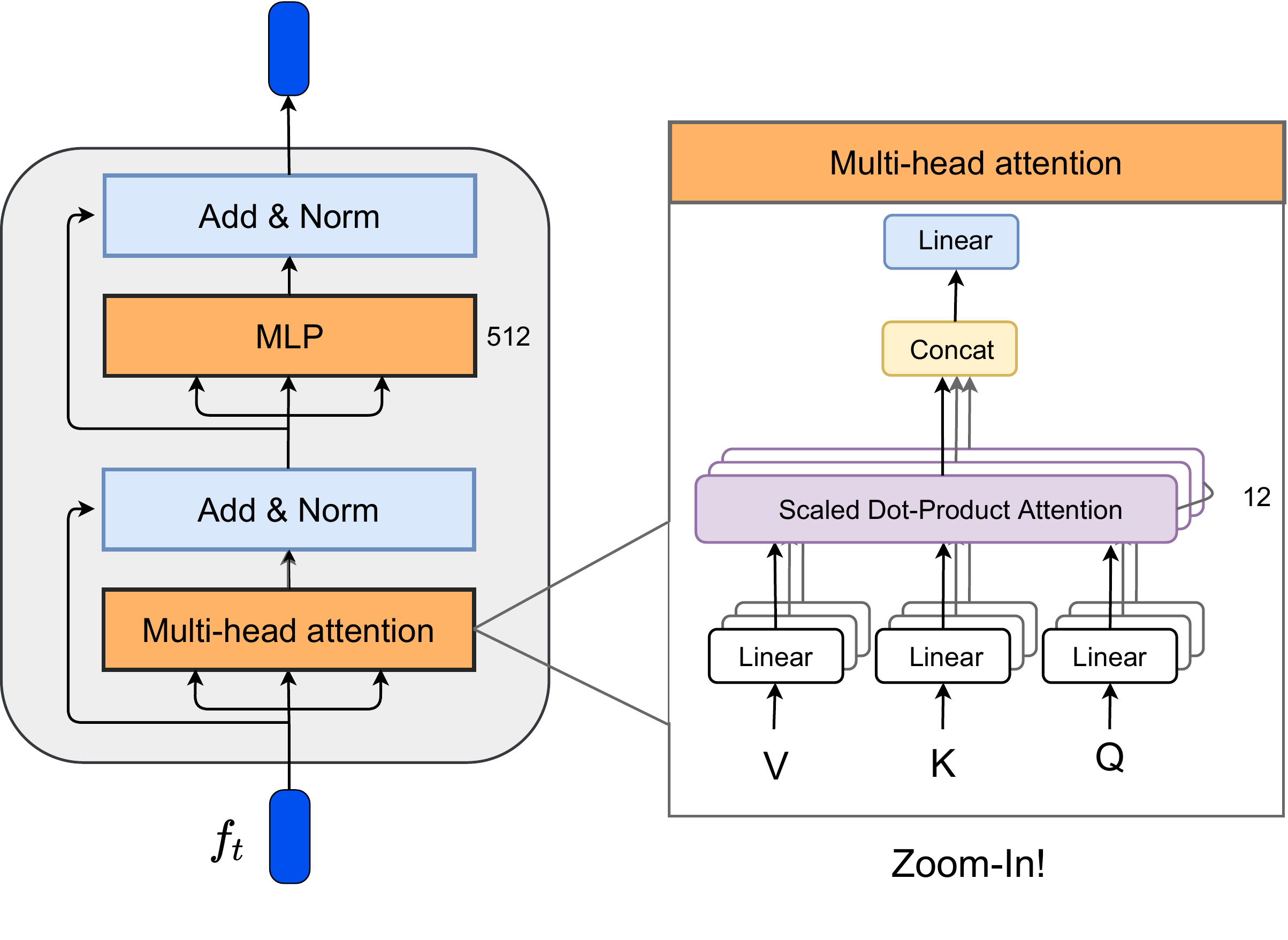}
	\vspace{-8mm}
	\caption{Architecture of our Transformer Encoder $\transformer$. }
	\label{fig:architecture_decoder}
\end{center}
\end{figure}
\begin{figure}[t]
\begin{center}
	\includegraphics[width=\linewidth, height=4cm]{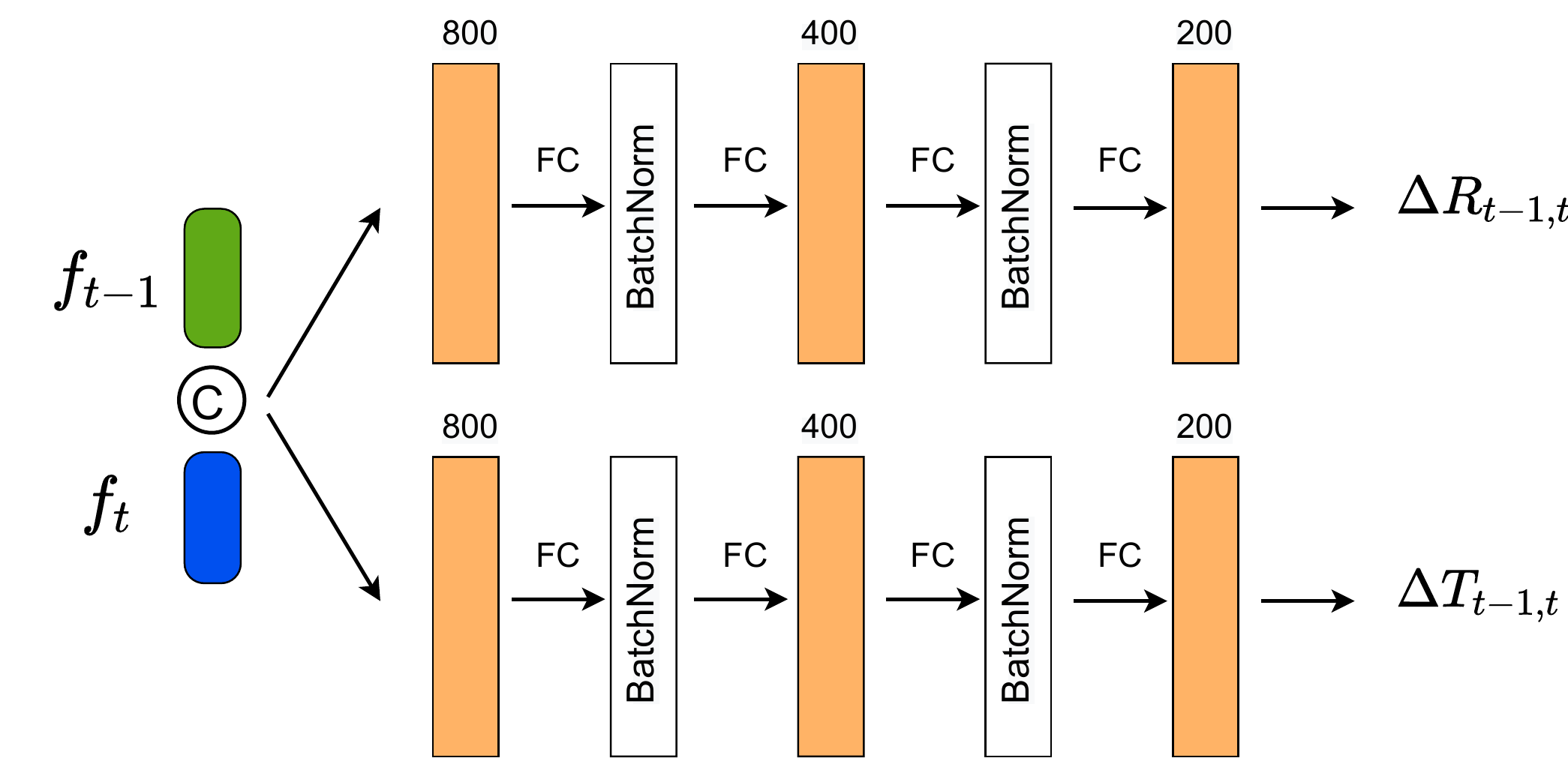}
	\vspace{-8mm}
	\caption{Architecture of our Pose Regressor $\regressor$. Given two global features $\feat_{t-1}$ and $\feat_{t}$ at the output of $\encoder$ in Our two-frame or of $\transformer$ in Our multi-frame, we concatenate them and predict relative pose change $\Delta \bR_{t-1,t}$ and $\Delta \bT_{t-1,t}$}
	\label{fig:architecture_decoder}
\end{center}
\end{figure}

\section{Additional results}

\subsection{Evaluation on Synthetic Datasets}

\begin{figure*}[t]
\begin{center}
\begin{tabular}{cc}
 \includegraphics[width=0.48\linewidth]{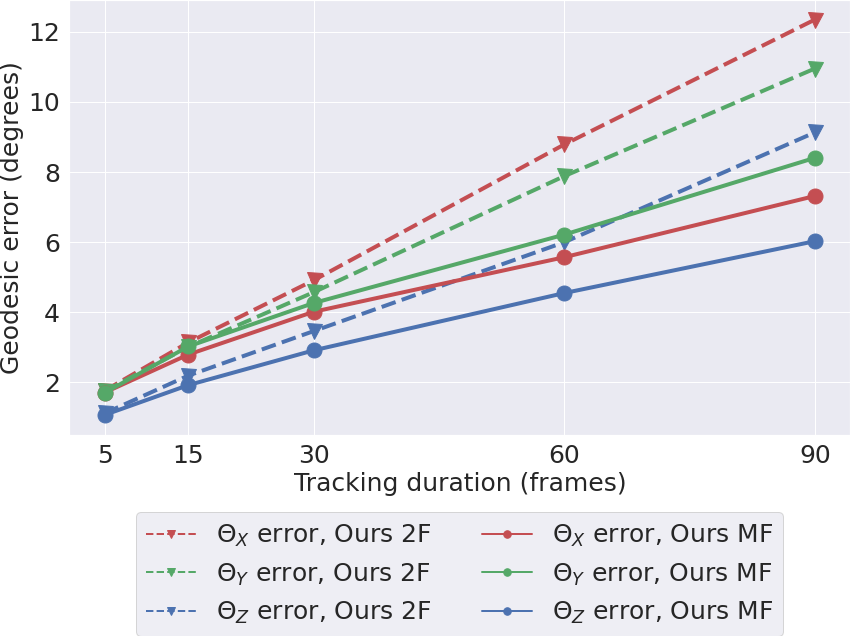} &
\includegraphics[width=0.48\linewidth]{images/supp/ablation_study_rotation_with_legend.png}
\end{tabular}
\end{center}
\vspace{-2mm}
\caption{{\bf Error accumulation over time for our two architectures.} \textbf{Left:} Rotation error over the 3 axes measured with Euler angles $(\theta_X, \theta_Y, \theta_Z)$; \textbf{Right:} Translation error over the 3 axes. Interestingly, the translation error along the X and Y axes increases much  slower with the multi-frame architecture than with the two-frame architecture. }
\label{fig:ablation_study_on_len_sequences}
\end{figure*}

\paragraph{Mitigating Error Accumulation.}
Tracking methods are prone to drift with time since they can accumulate of errors between each pair of consecutive frames. Figure~\ref{fig:ablation_study_on_len_sequences} shows the average of rotation and translation errors over time on the test set of our ShapeNet  dataset. As can be seen from the graphs, the multi-frame Transformer-based architecture. accumulates significantly less errors, making it a good candidate for tracking applications.

\paragraph{Tracking translation in Z axis.}
As discussed in Section~4.3 of the main paper, most translation errors come from the Z camera axis for both methods when the tracking duration is small. This is because estimating motion along this axis is notoriously very challenging for RGB-based methods, as relatively large $\Delta T_z$ can occur even with minimal appearance change between two frames.\\

\vspace{-2mm}
\noindent We show additional qualitative results on the ModelNet dataset in Figure~\ref{fig:modelnet_qualitative_results}. 

\subsection{Evaluation on Real-world Datasets}

\begin{figure}[!t]
\begin{center}
\begin{tabular}{c}
 \includegraphics[width=0.97\linewidth]{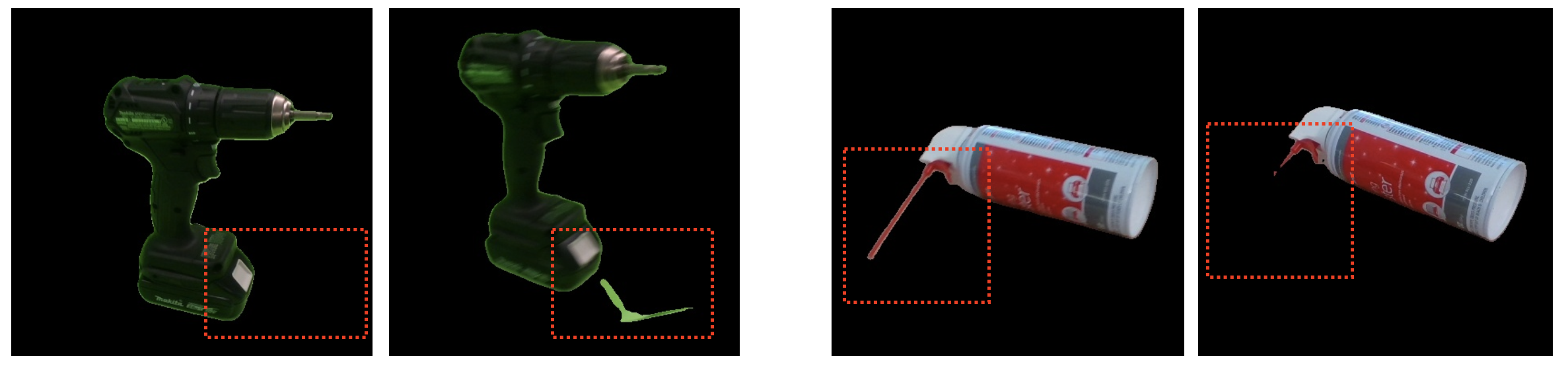}
\end{tabular}
\end{center}
\vspace{-5mm}
\caption{\textbf{Failure case of segmentation model \cite{Du20211stPS}} in the highlight region in \textbf{\color{red}red} color which creates “motion ambiguities" as the appearance changes are explained by other motions than correct ones.}
\label{fig:failure_case_segmentation_model}
\end{figure}

\paragraph{Failure cases of unseen object segmentation.} We show in Figure~\ref{fig:failure_case_segmentation_model} some failure cases of the segmentation model~\cite{Du20211stPS} on unseen objects of Laval~\cite{garon2018framework} and Moped~\cite{park2020latent}.

\vspace{1mm}
\begin{figure*}[t]
\setlength\qualmodelnetheight{2.5cm}
\begin{center}
\begin{tabular}{cccccc}
 \includegraphics[height=\qualmodelnetheight, width=\qualmodelnetheight]{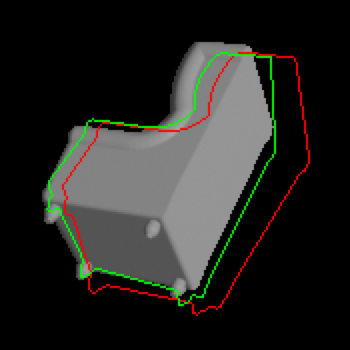} &
 \includegraphics[ height=\qualmodelnetheight, width=\qualmodelnetheight]{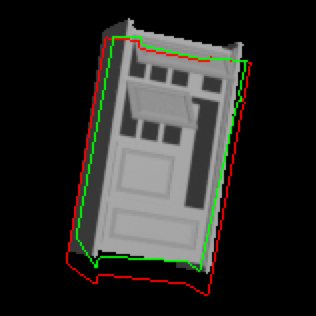} &
 \includegraphics[ height=\qualmodelnetheight, width=\qualmodelnetheight]{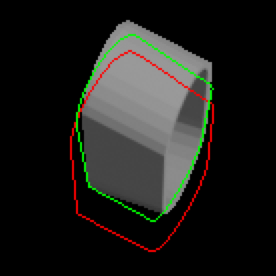} &
 \includegraphics[ height=\qualmodelnetheight, width=\qualmodelnetheight]{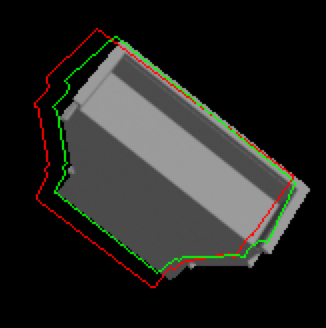}&
 \includegraphics[ height=\qualmodelnetheight, width=\qualmodelnetheight]{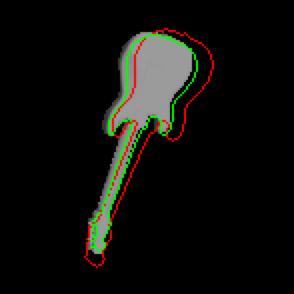} &
 \includegraphics[ height=\qualmodelnetheight, width=\qualmodelnetheight]{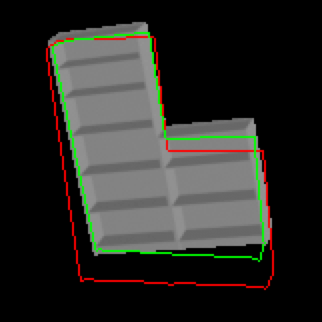}
\end{tabular}\\
\begin{tabular}{cccccc}
 \includegraphics[height=\qualmodelnetheight, width=\qualmodelnetheight]{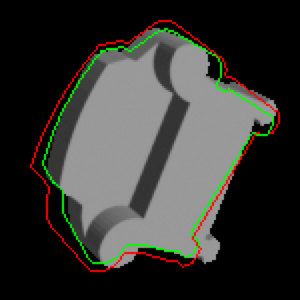} &
 \includegraphics[ height=\qualmodelnetheight, width=\qualmodelnetheight]{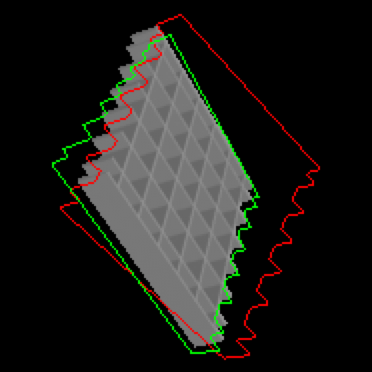} &
 \includegraphics[ height=\qualmodelnetheight, width=\qualmodelnetheight]{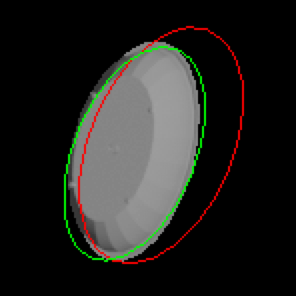} &
 \includegraphics[ height=\qualmodelnetheight, width=\qualmodelnetheight]{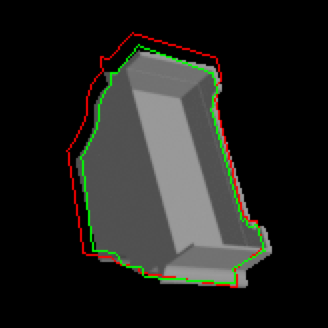}&
 \includegraphics[ height=\qualmodelnetheight, width=\qualmodelnetheight]{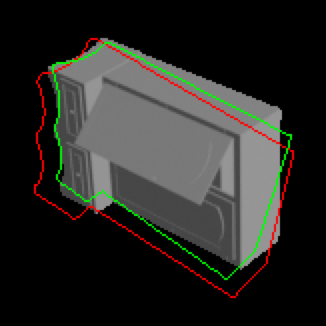} &
 \includegraphics[ height=\qualmodelnetheight, width=\qualmodelnetheight]{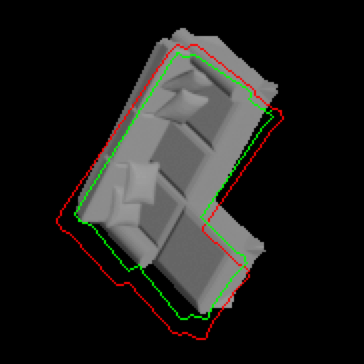}
\end{tabular}\\
\begin{tabular}{cccccc}
 \includegraphics[height=\qualmodelnetheight, width=\qualmodelnetheight]{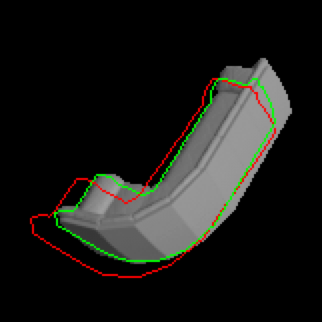} &
 \includegraphics[ height=\qualmodelnetheight, width=\qualmodelnetheight]{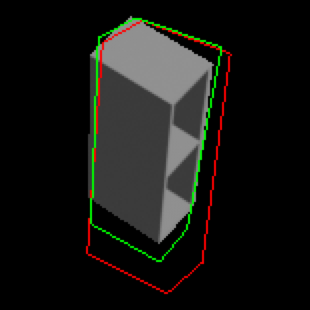} &
 \includegraphics[ height=\qualmodelnetheight, width=\qualmodelnetheight]{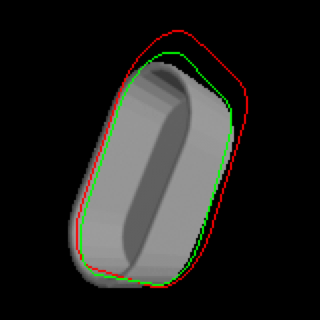} &
 \includegraphics[ height=\qualmodelnetheight, width=\qualmodelnetheight]{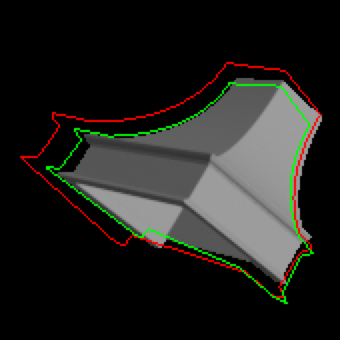}&
 \includegraphics[ height=\qualmodelnetheight, width=\qualmodelnetheight]{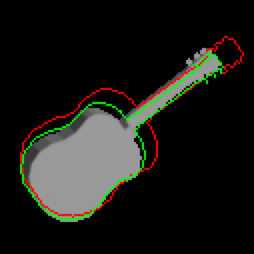} &
 \includegraphics[ height=\qualmodelnetheight, width=\qualmodelnetheight]{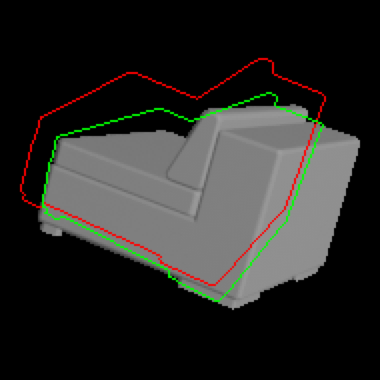}
\end{tabular}\\
\end{center}
\vspace{-2mm}
\caption{\textbf{Qualitative results on ModelNet.} We show the target image with an initial perturbed pose in a \textbf{\color{red}red} wireframe and our pose prediction with a \textbf{\color{green}green} wireframes. The object wireframes are used only for visualization.}
\label{fig:modelnet_qualitative_results}
\end{figure*}

\begin{figure*}[h]
\begin{center}
	\includegraphics[width=0.90\linewidth]{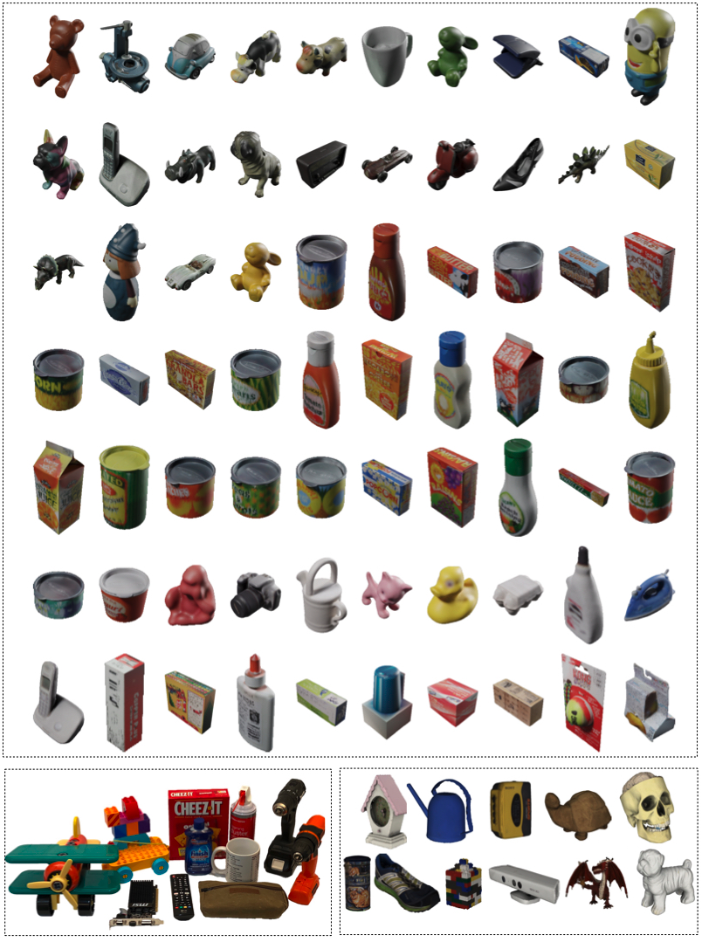}
	\vspace{-4mm}
	\caption{Visual comparison between training objects from BOP challenges \cite{hodan2018bop} (top) and testing objects of Moped \cite{park2020latent} (bottom left) and Laval \cite{garon2018framework} datasets (bottom right). The bottom images are taken from \cite{park2020latent,garon2018framework}.}\label{fig:visualize_objects}
\end{center}
\end{figure*}

\paragraph{Training objects from BOP \cite{hodan2018bop}.} We train our approach on a dataset made only of synthetic data created from rendering CAD models from BOP challenges \cite{hodan2018bop} including: LINEMOD~\cite{Hinterstoier2012ModelBT}, HB~\cite{kaskman2019homebreweddb}, HOPE, RU-APC~\cite{rennie2016dataset} datasets with BlenderProc~\cite{denninger2019blenderproc}. We visually compare the training and testing objects in Figure~\ref{fig:visualize_objects}.\\

\begin{figure*}[t]
\setlength\quallavalheight{2.2cm}
\setlength\qualmopoedheight{2.3cm}
\begin{center}
\begin{tabular}{cccccc}
 \includegraphics[ height=\quallavalheight]{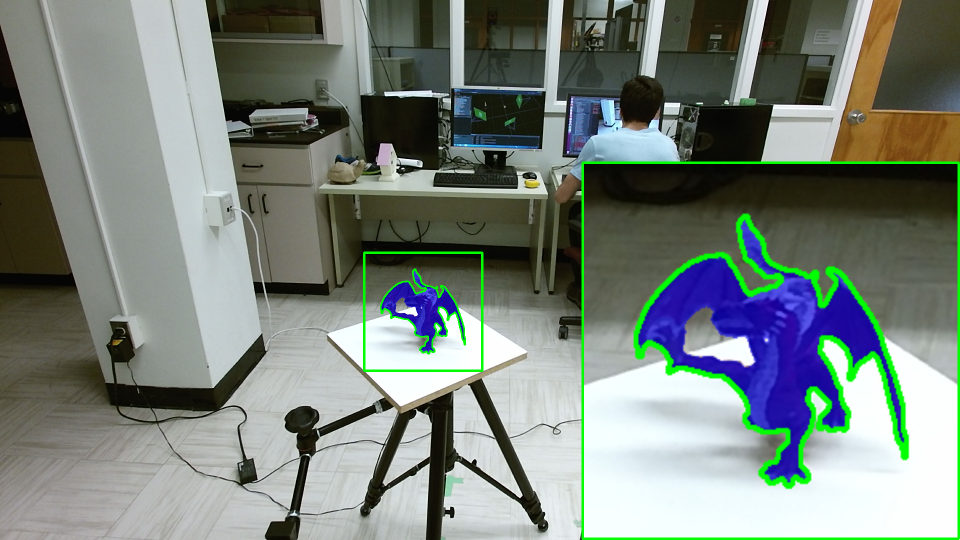} &
 \includegraphics[ height=\quallavalheight]{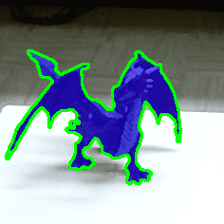} &
 \includegraphics[ height=\quallavalheight]{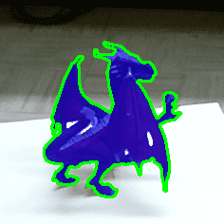} &
 \includegraphics[ height=\quallavalheight]{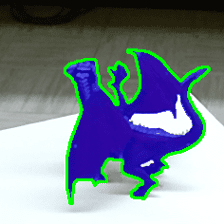}&
 \includegraphics[ height=\quallavalheight]{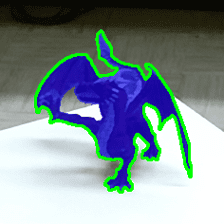}&
 \includegraphics[ height=\quallavalheight]{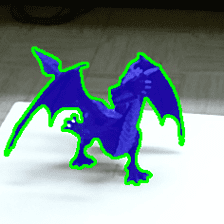}
\end{tabular}\\
\begin{tabular}{cccccc}
 \includegraphics[ height=\quallavalheight]{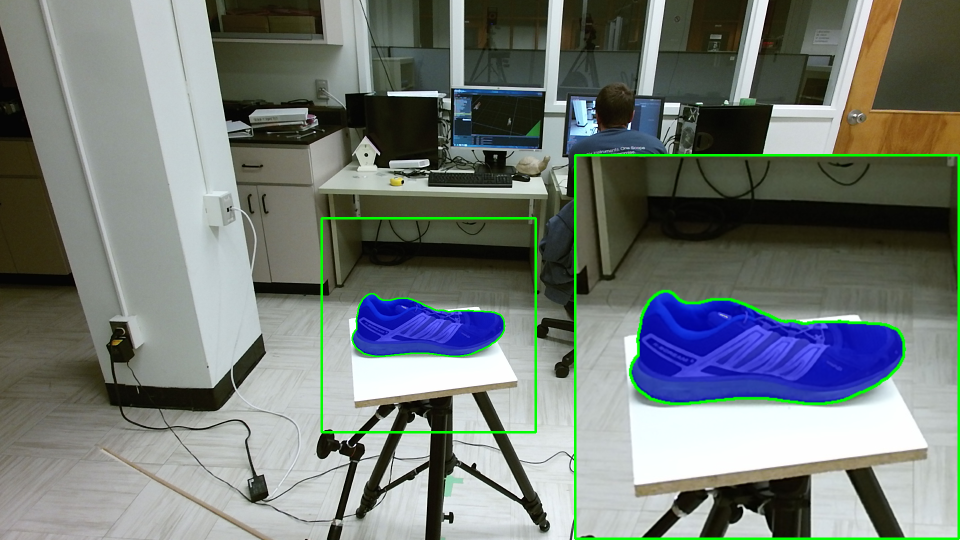}&
 \includegraphics[ height=\quallavalheight]{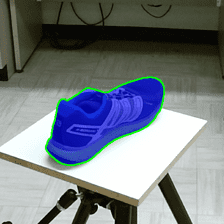} &
 \includegraphics[ height=\quallavalheight]{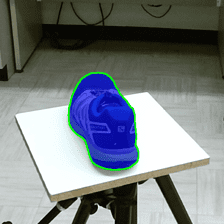} &
 \includegraphics[ height=\quallavalheight]{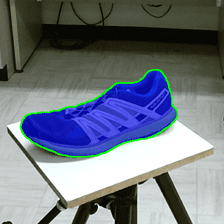}&
 \includegraphics[ height=\quallavalheight]{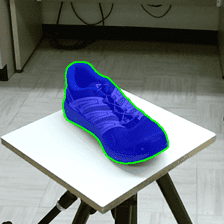}&
 \includegraphics[ height=\quallavalheight]{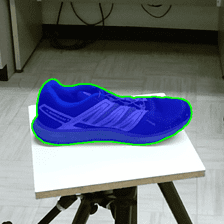}
\end{tabular}\\
\begin{tabular}{cccccc}
 \includegraphics[ height=\quallavalheight]{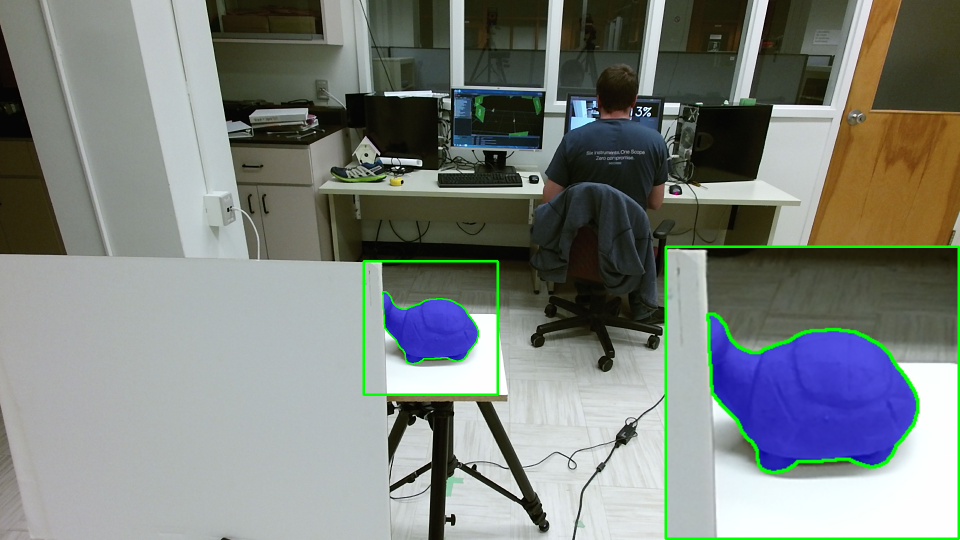}&
 \includegraphics[ height=\quallavalheight]{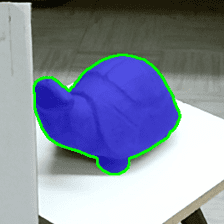} &
 \includegraphics[ height=\quallavalheight]{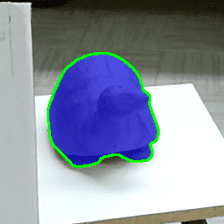} &
 \includegraphics[ height=\quallavalheight]{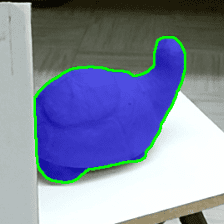}&
 \includegraphics[ height=\quallavalheight]{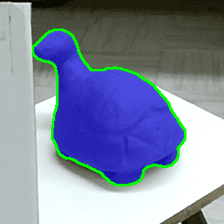}&
 \includegraphics[ height=\quallavalheight]{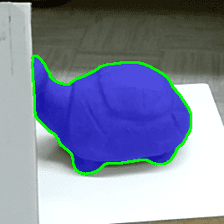}
\end{tabular}\\
\begin{tabular}{cccccc}
 \includegraphics[ height=\quallavalheight]{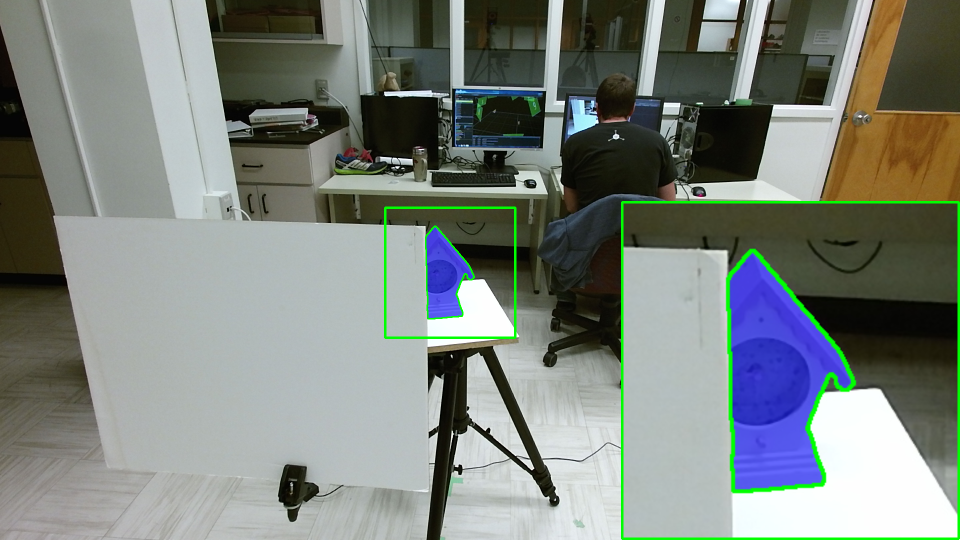}&
 \includegraphics[ height=\quallavalheight]{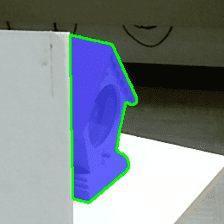} &
 \includegraphics[ height=\quallavalheight]{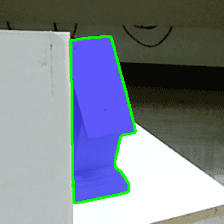} &
 \includegraphics[ height=\quallavalheight]{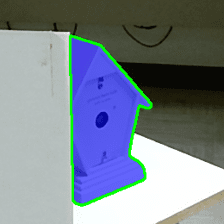}&
 \includegraphics[ height=\quallavalheight]{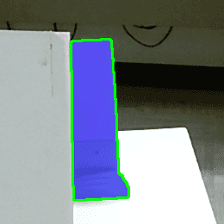}&
 \includegraphics[ height=\quallavalheight]{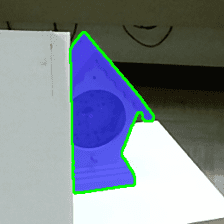}
\end{tabular}\\
\begin{tabular}{ccccc}
 \includegraphics[ height=\qualmopoedheight]{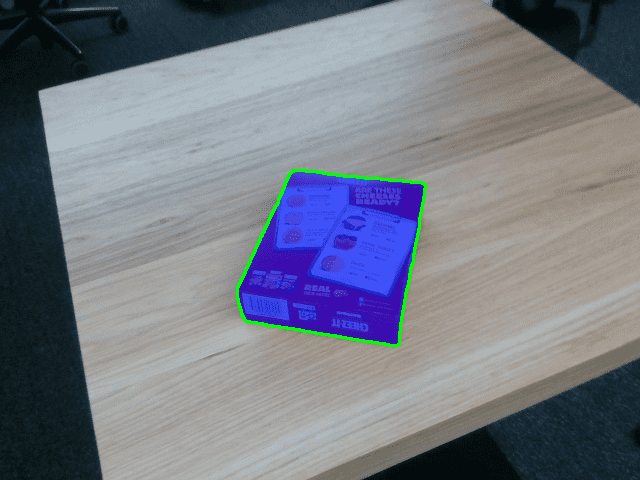} &
 \includegraphics[ height=\qualmopoedheight]{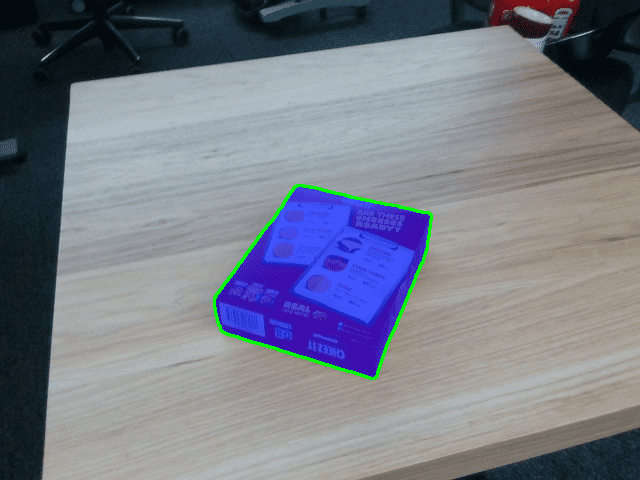} &
 \includegraphics[ height=\qualmopoedheight]{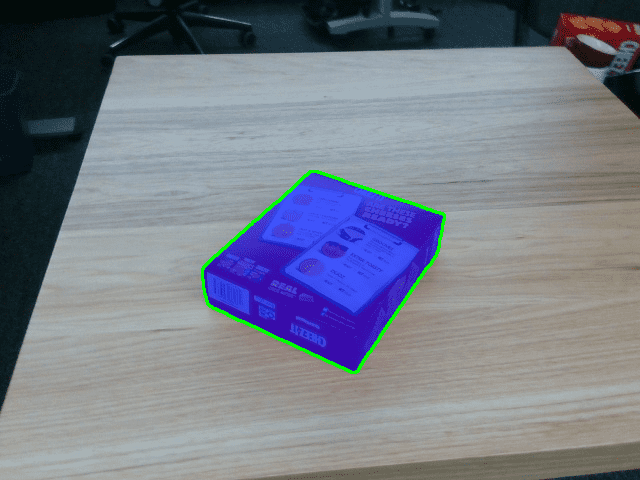} &
 \includegraphics[ height=\qualmopoedheight]{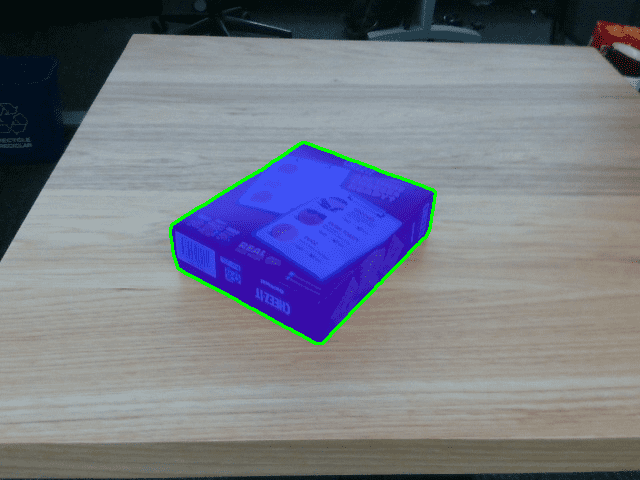}&
 \includegraphics[ height=\qualmopoedheight]{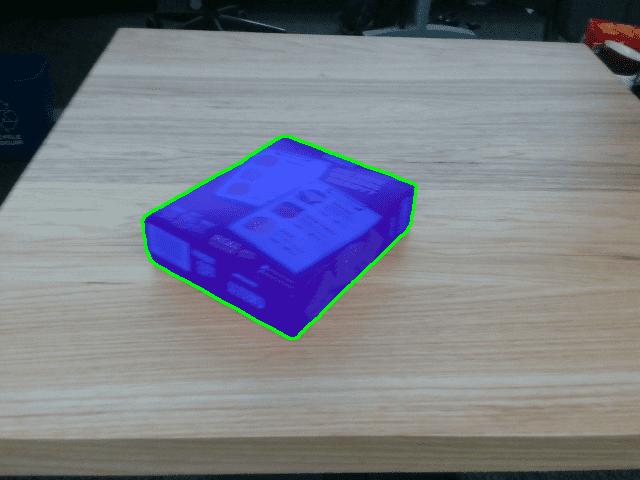}
\end{tabular}\\
\begin{tabular}{ccccc}
 \includegraphics[height=\qualmopoedheight]{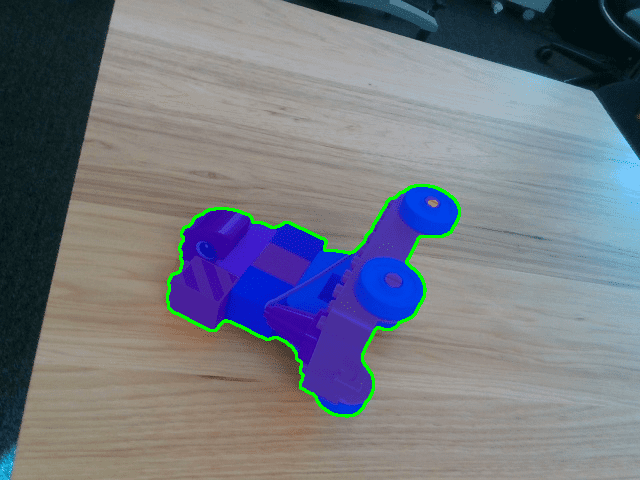} &
 \includegraphics[height=\qualmopoedheight]{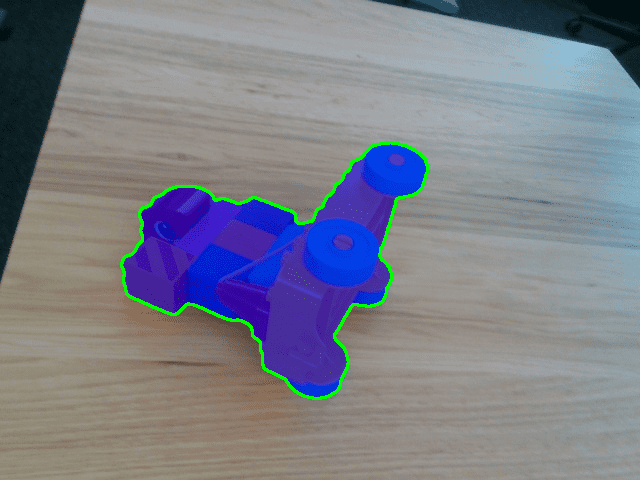} &
 \includegraphics[height=\qualmopoedheight]{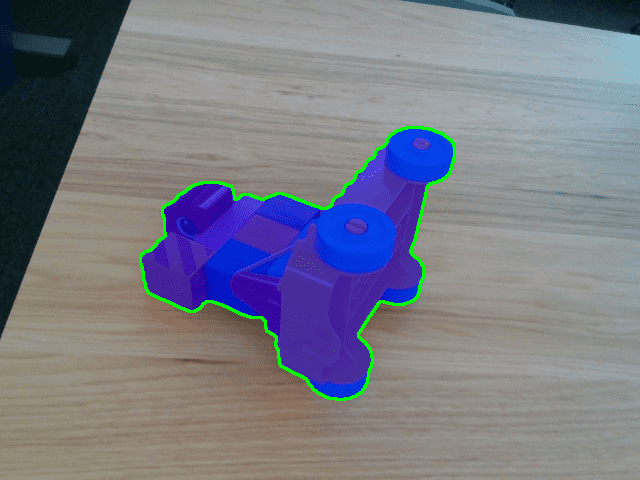} &
 \includegraphics[height=\qualmopoedheight]{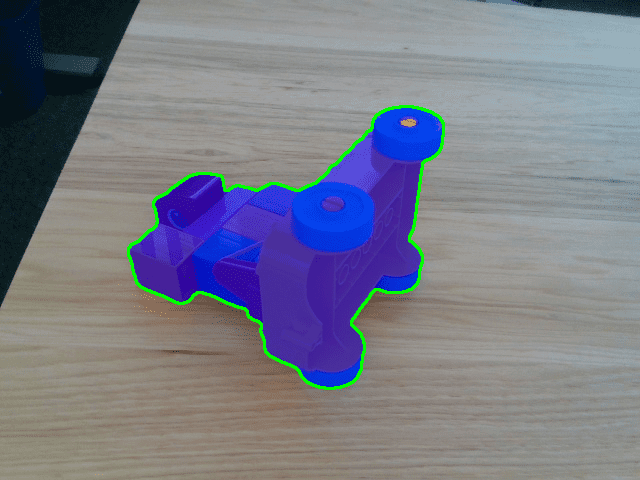}&
 \includegraphics[height=\qualmopoedheight]{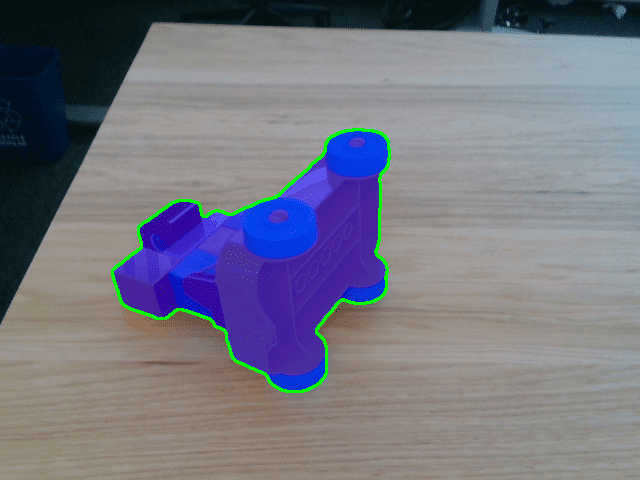}
\end{tabular}\\
\begin{tabular}{ccccc}
 \includegraphics[ height=\qualmopoedheight]{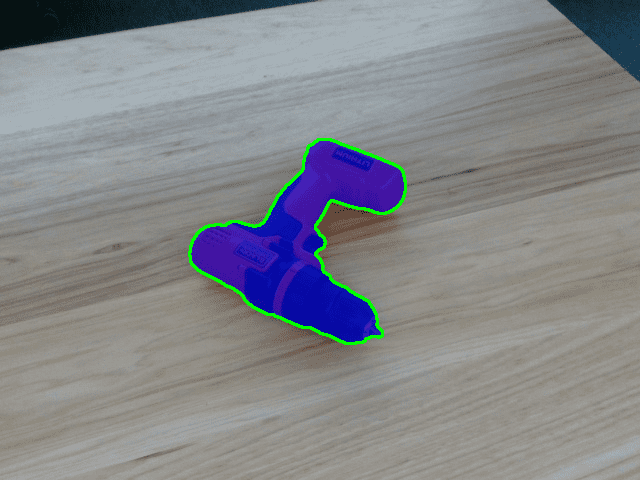} &
 \includegraphics[ height=\qualmopoedheight]{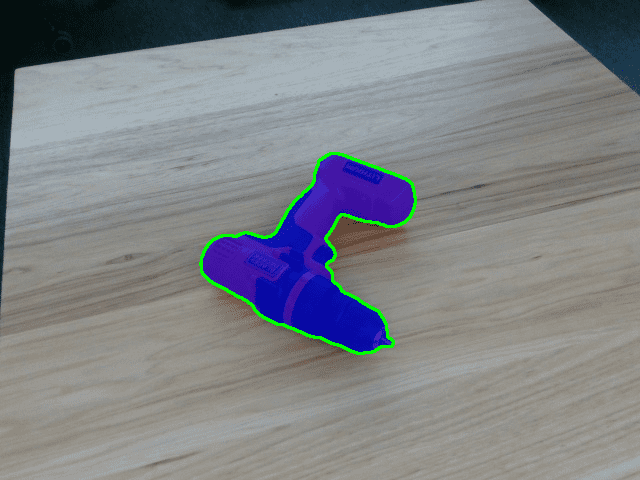} &
 \includegraphics[ height=\qualmopoedheight]{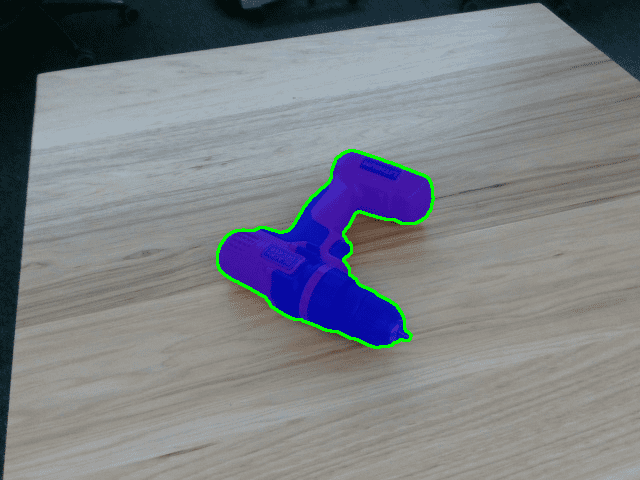} &
 \includegraphics[ height=\qualmopoedheight]{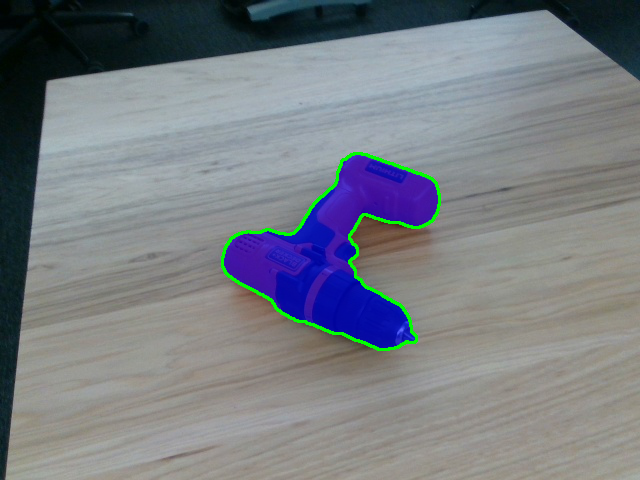}&
 \includegraphics[ height=\qualmopoedheight]{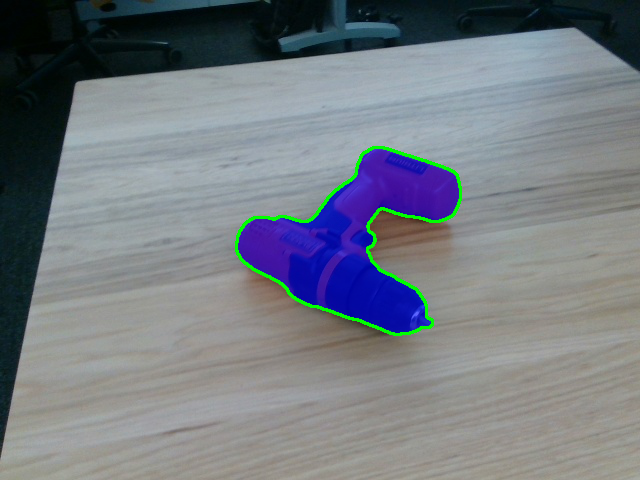}
\end{tabular}\\
\begin{tabular}{ccccc}
 \includegraphics[ height=\qualmopoedheight]{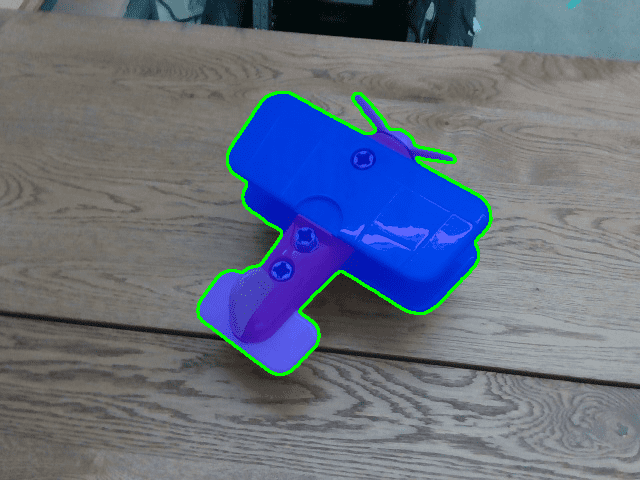} &
 \includegraphics[ height=\qualmopoedheight]{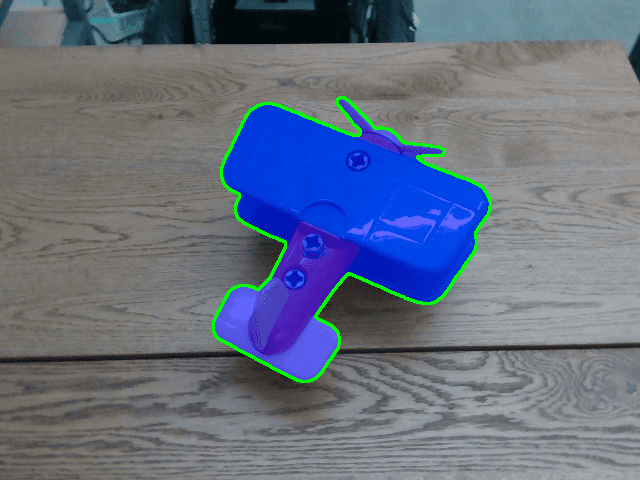} &
 \includegraphics[ height=\qualmopoedheight]{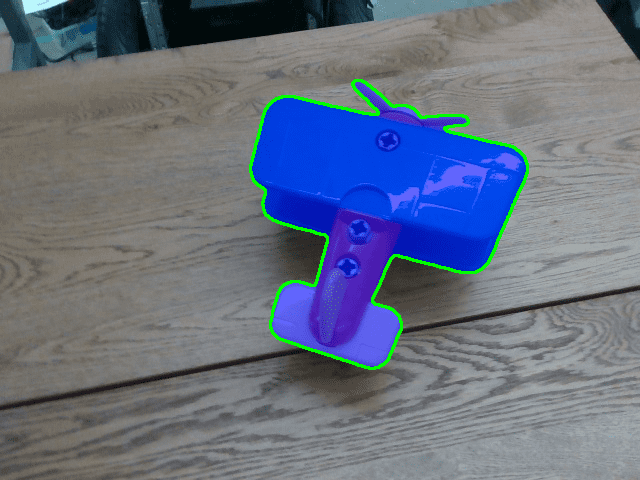} &
 \includegraphics[ height=\qualmopoedheight]{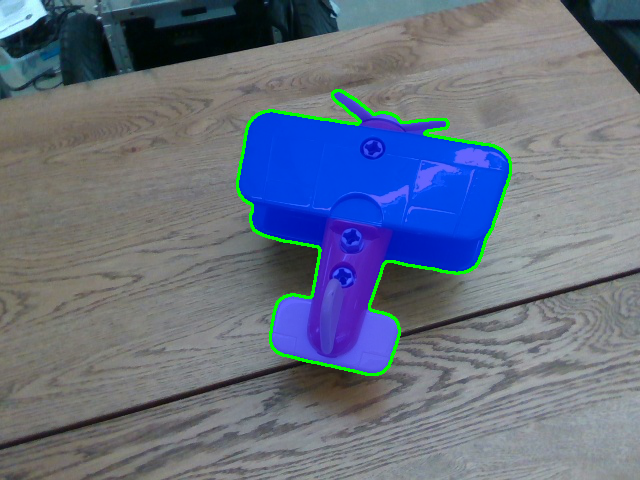}&
 \includegraphics[ height=\qualmopoedheight]{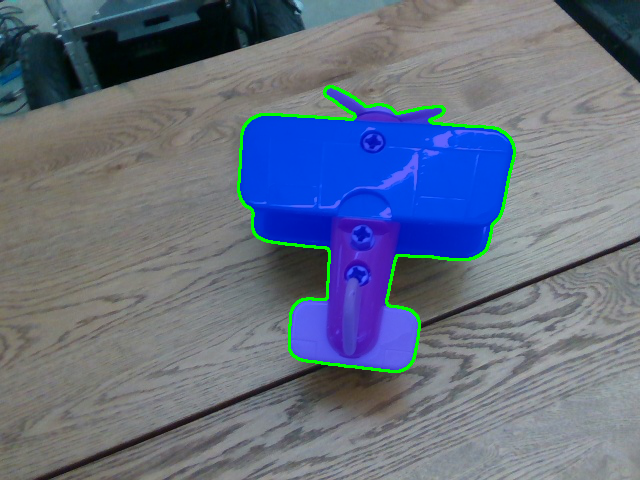}
\end{tabular}\\
\begin{tabular}{ccccc}
 \includegraphics[ height=\qualmopoedheight]{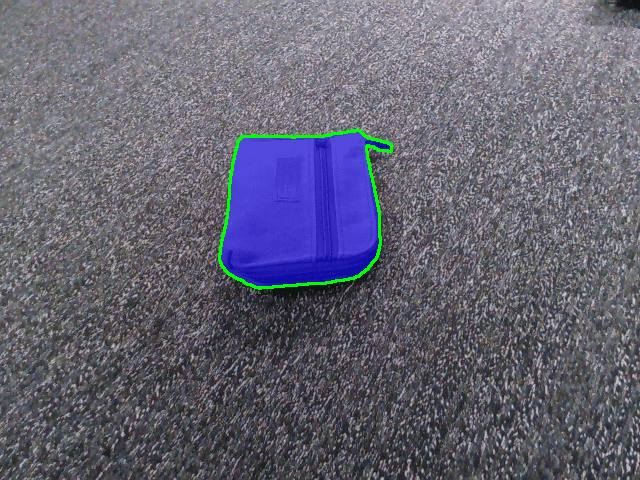}&
 \includegraphics[ height=\qualmopoedheight]{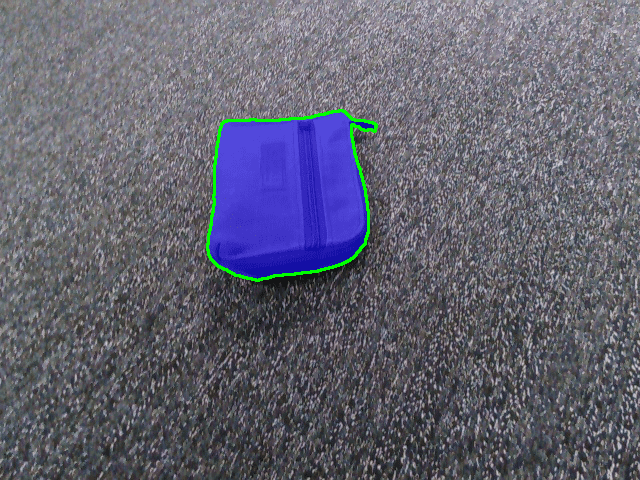} &
 \includegraphics[ height=\qualmopoedheight]{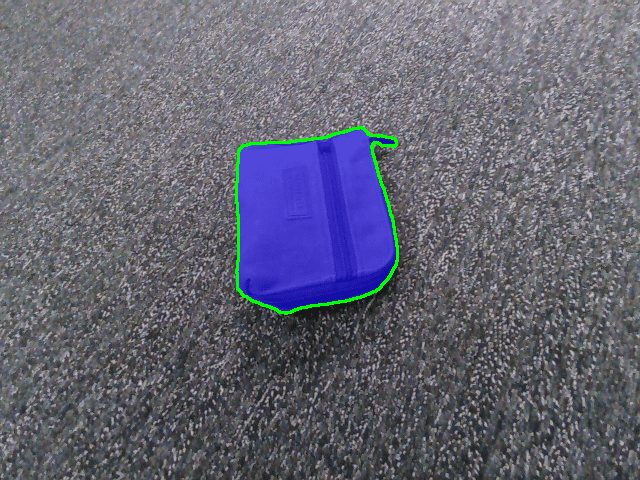} &
 \includegraphics[ height=\qualmopoedheight]{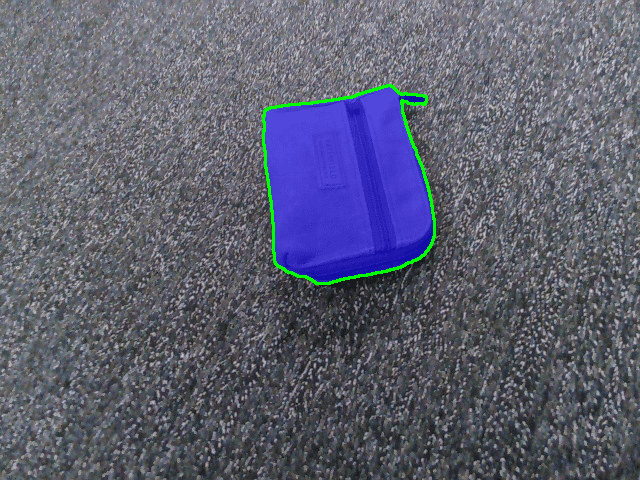}&
 \includegraphics[ height=\qualmopoedheight]{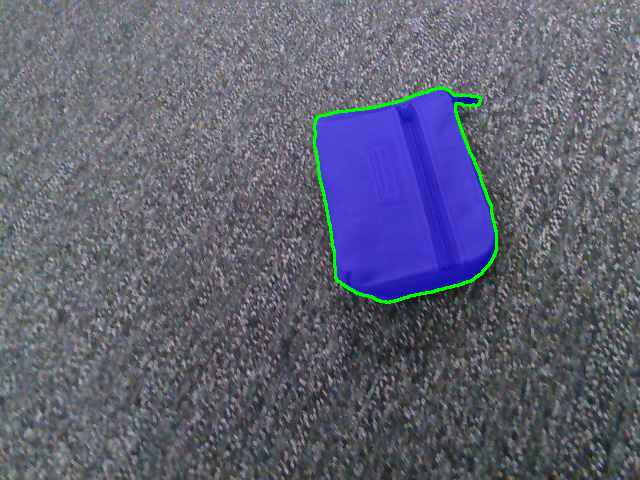}
\end{tabular}\\
\begin{tabular}{@{\;\;\;\;\;\;\;\;\;\;\;}c}
\vspace{3mm}
\includegraphics[height=0.20\qualmopoedheight]{images/introduction/time_arrow.pdf}
\end{tabular}
\end{center}
\vspace{-10mm}
\caption{\textbf{The segmentation model of \cite{Du20211stPS} works surprisingly well on unseen objects. } Results on Laval and Moped images. }
\label{fig:segmentation_results}
\end{figure*}

\begin{figure*}[t]
\setlength\quallavalheight{2.2cm}
\setlength\qualmopoedheight{2.3cm}
\begin{center}
\begin{tabular}{cccccc}
 \includegraphics[ height=\quallavalheight]{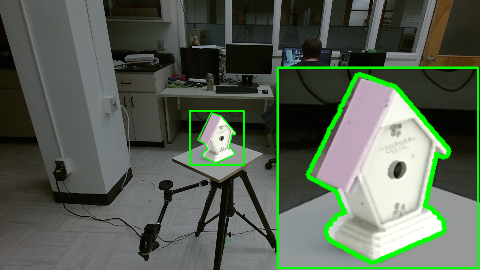} &
 \includegraphics[ height=\quallavalheight, width=\quallavalheight]{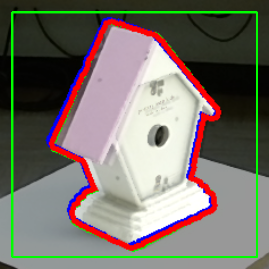} &
 \includegraphics[ height=\quallavalheight, width=\quallavalheight]{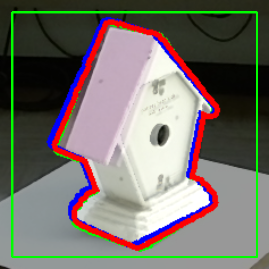} &
 \includegraphics[ height=\quallavalheight, width=\quallavalheight]{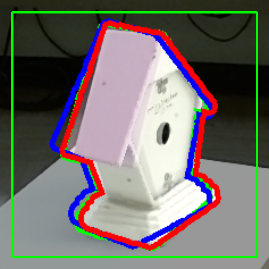}&
 \includegraphics[ height=\quallavalheight, width=\quallavalheight]{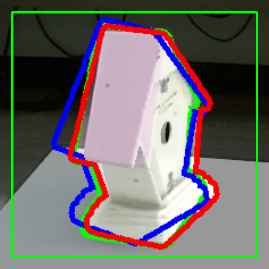}&
 \includegraphics[ height=\quallavalheight, width=\quallavalheight]{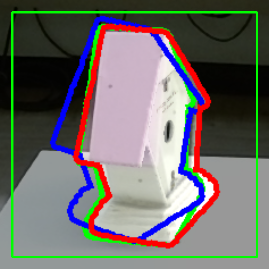}
\end{tabular}\\
\begin{tabular}{cccccc}
 \includegraphics[ height=\quallavalheight]{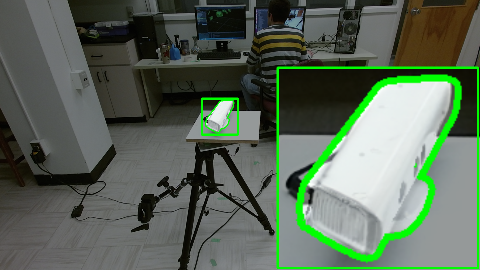}&
 \includegraphics[ height=\quallavalheight, width=\quallavalheight]{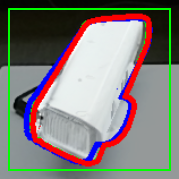} &
 \includegraphics[ height=\quallavalheight, width=\quallavalheight]{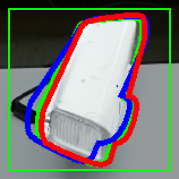} &
 \includegraphics[ height=\quallavalheight, width=\quallavalheight]{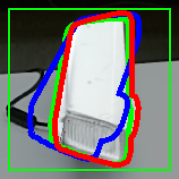}&
 \includegraphics[ height=\quallavalheight, width=\quallavalheight]{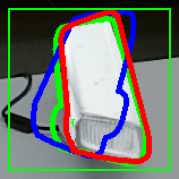}&
 \includegraphics[ height=\quallavalheight, width=\quallavalheight]{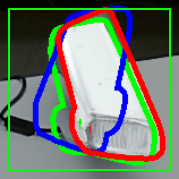}
\end{tabular}\\
\begin{tabular}{ccccc}
 \includegraphics[ height=\qualmopoedheight]{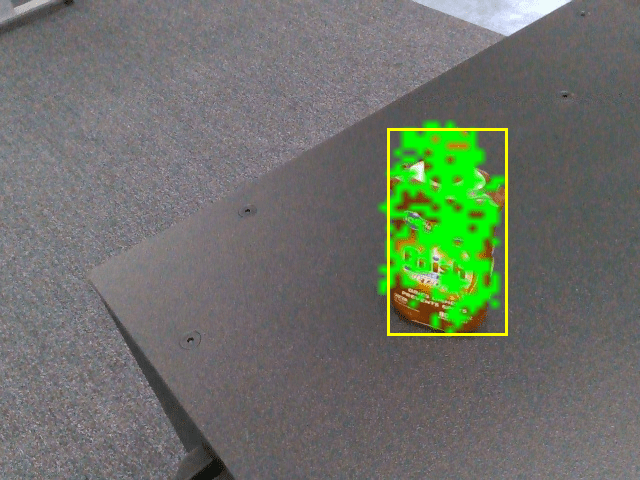} &
 \includegraphics[ height=\qualmopoedheight]{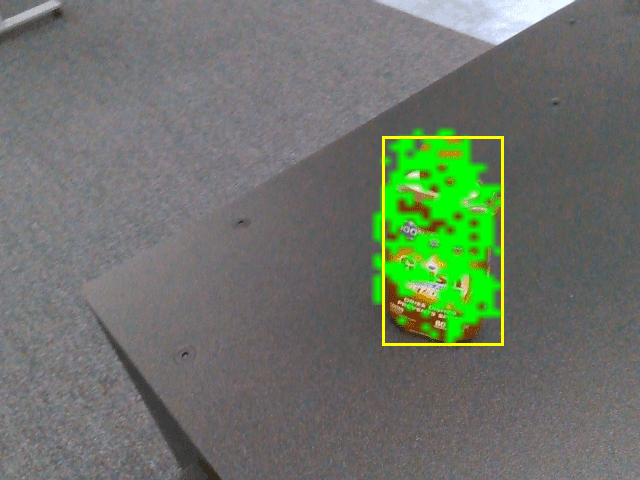} &
 \includegraphics[ height=\qualmopoedheight]{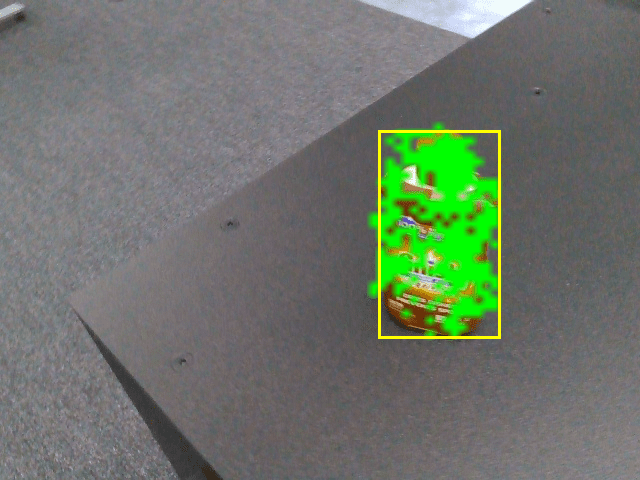} &
 \includegraphics[ height=\qualmopoedheight]{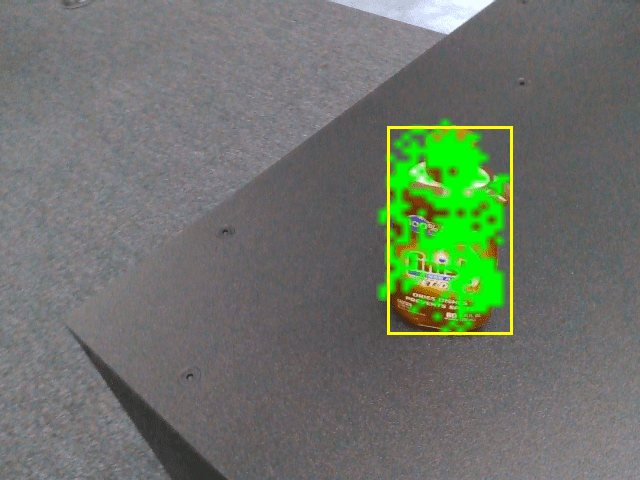}&
 \includegraphics[ height=\qualmopoedheight]{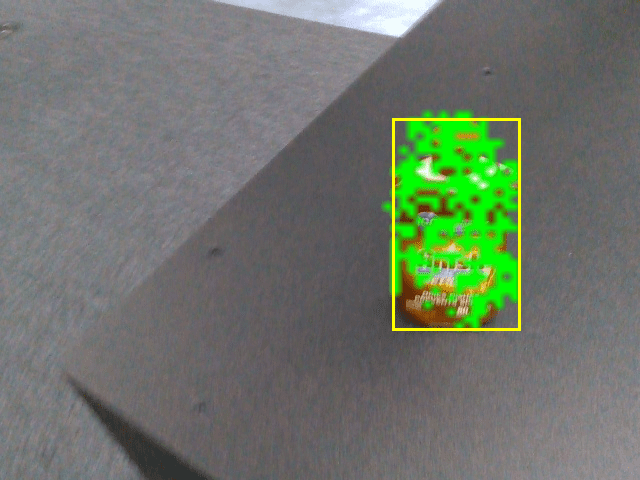}
\end{tabular}\\
\begin{tabular}{ccccc}
 \includegraphics[ height=\qualmopoedheight]{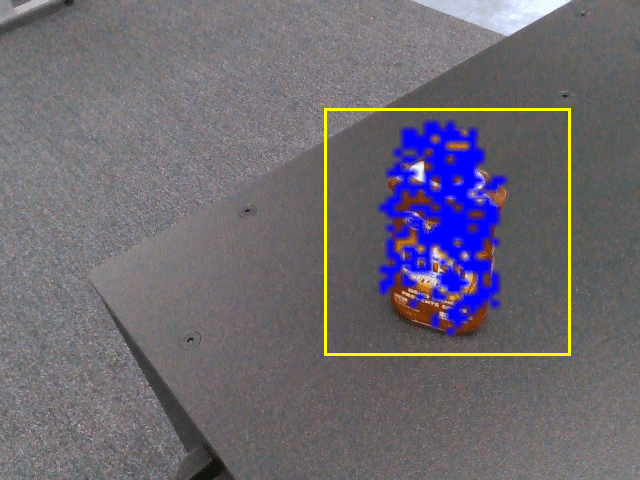} &
 \includegraphics[ height=\qualmopoedheight]{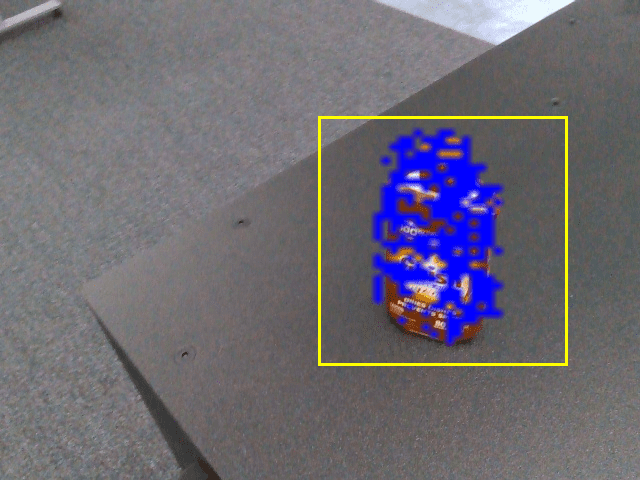} &
 \includegraphics[ height=\qualmopoedheight]{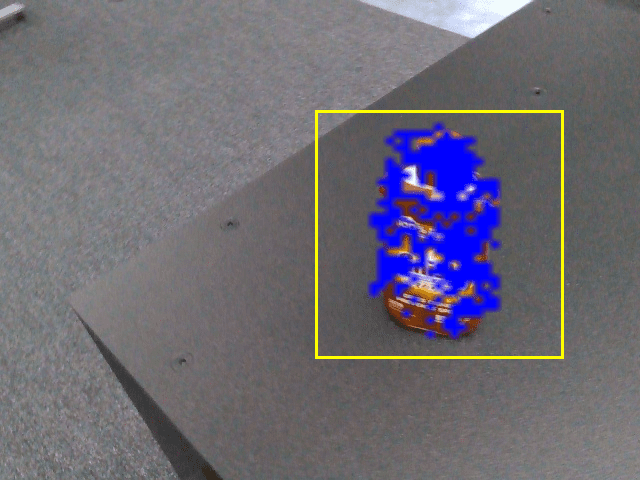} &
 \includegraphics[ height=\qualmopoedheight]{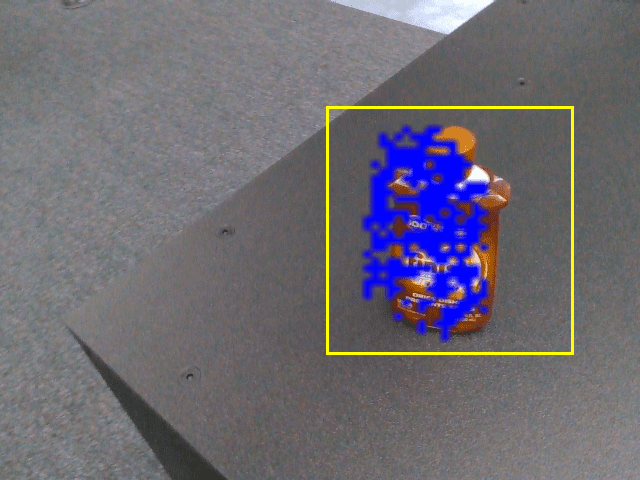}&
 \includegraphics[ height=\qualmopoedheight]{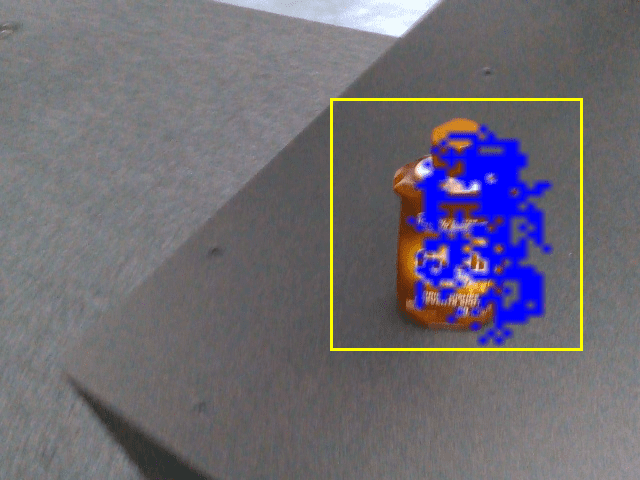}
\end{tabular}\\
\begin{tabular}{ccccc}
 \includegraphics[ height=\qualmopoedheight]{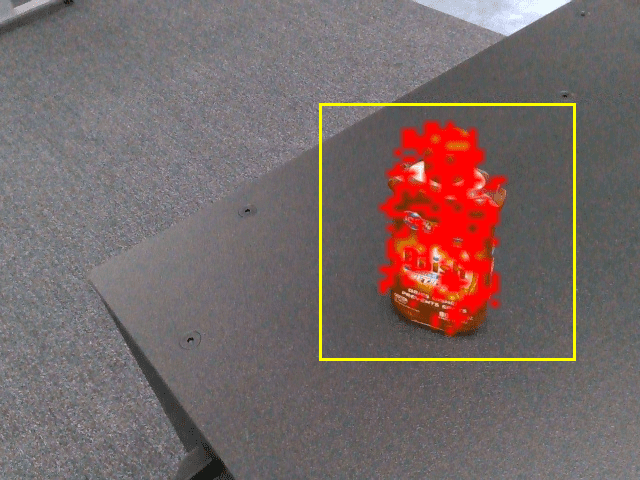} &
 \includegraphics[ height=\qualmopoedheight]{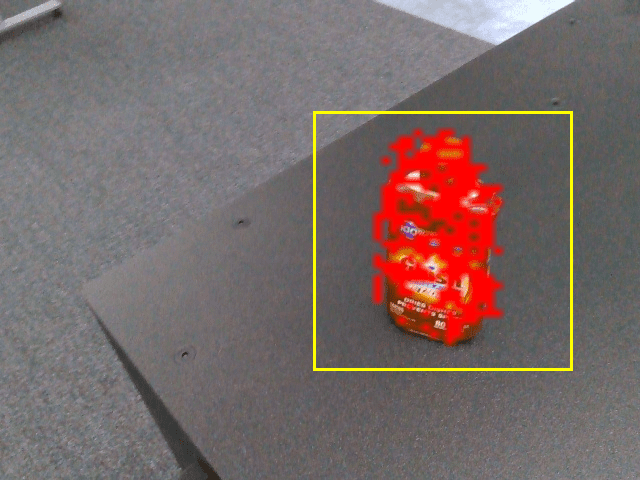} &
 \includegraphics[ height=\qualmopoedheight]{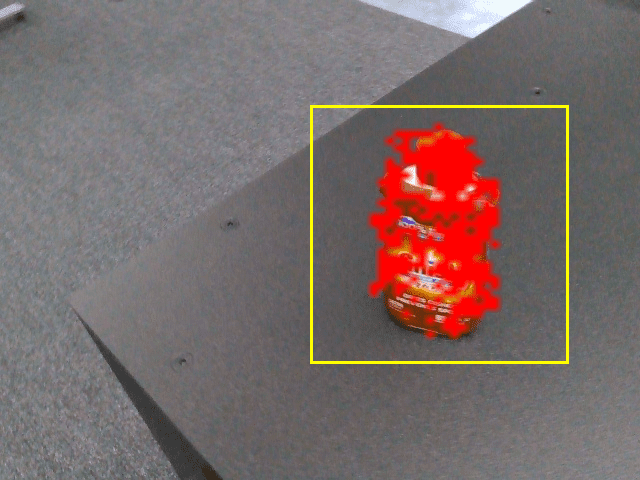} &
 \includegraphics[ height=\qualmopoedheight]{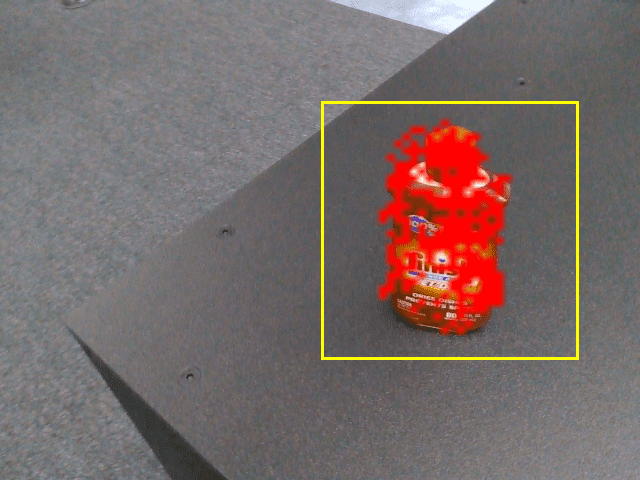}&
 \includegraphics[ height=\qualmopoedheight]{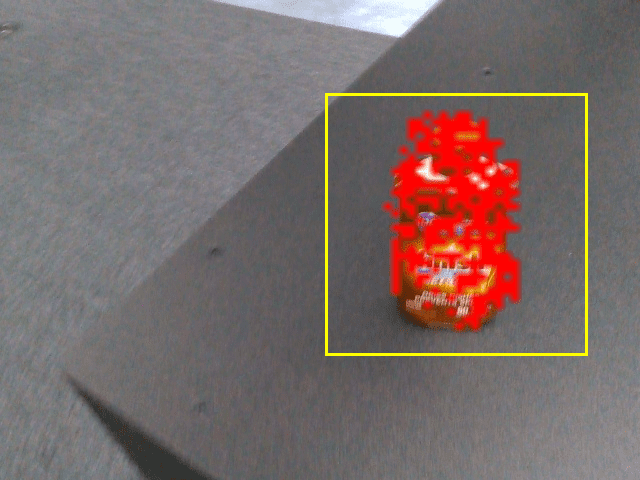}
\end{tabular}\\
\begin{tabular}{ccccc}
 \includegraphics[height=\qualmopoedheight]{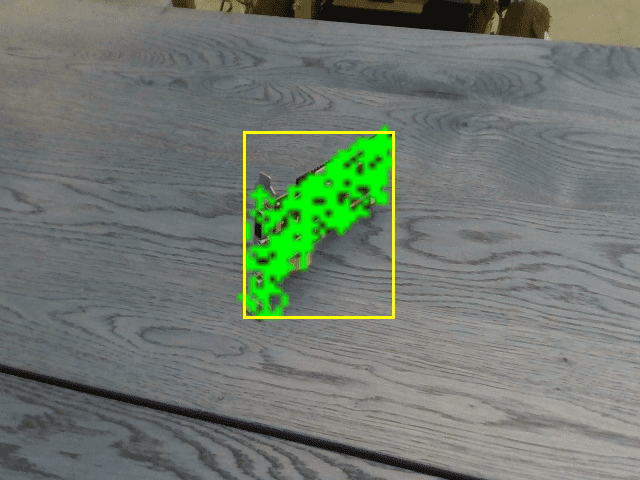} &
 \includegraphics[height=\qualmopoedheight]{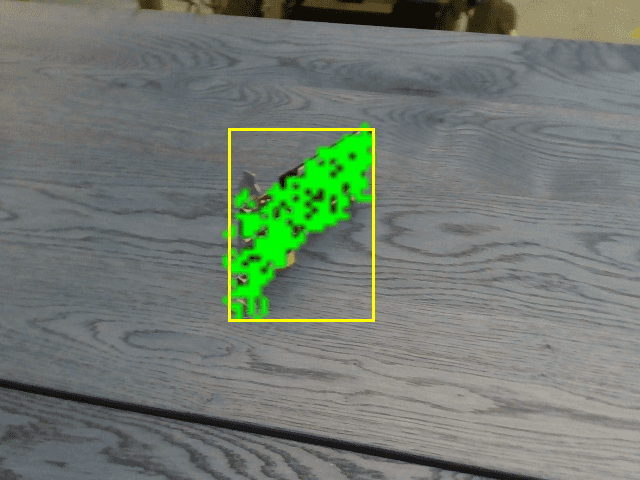} &
 \includegraphics[height=\qualmopoedheight]{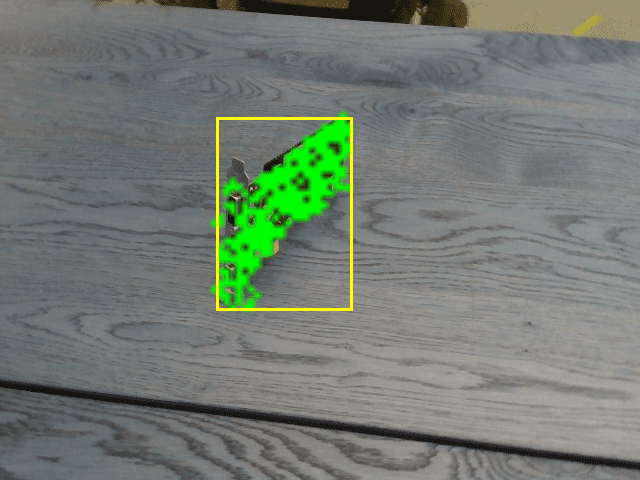} &
 \includegraphics[height=\qualmopoedheight]{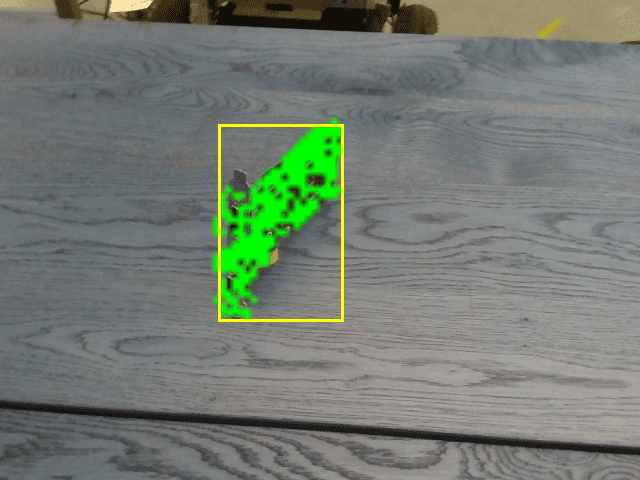}&
 \includegraphics[height=\qualmopoedheight]{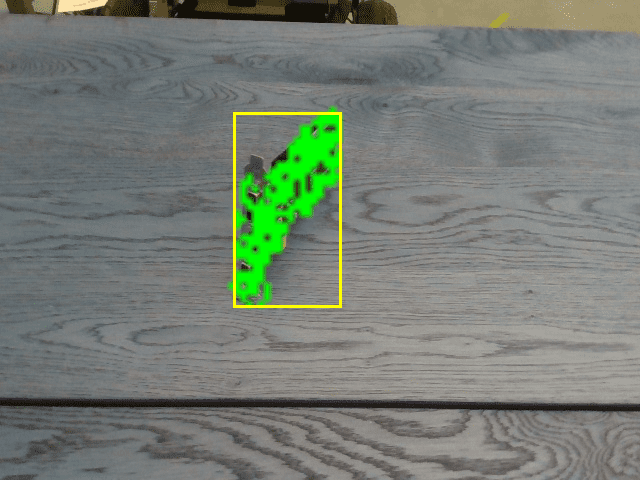}
\end{tabular}\\
\begin{tabular}{ccccc}
 \includegraphics[height=\qualmopoedheight]{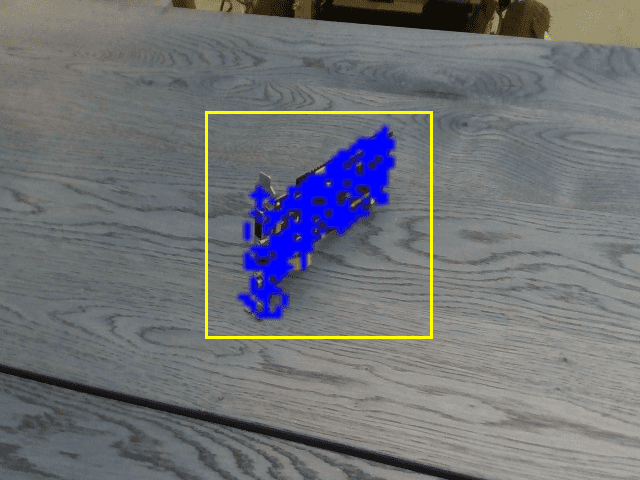} &
 \includegraphics[height=\qualmopoedheight]{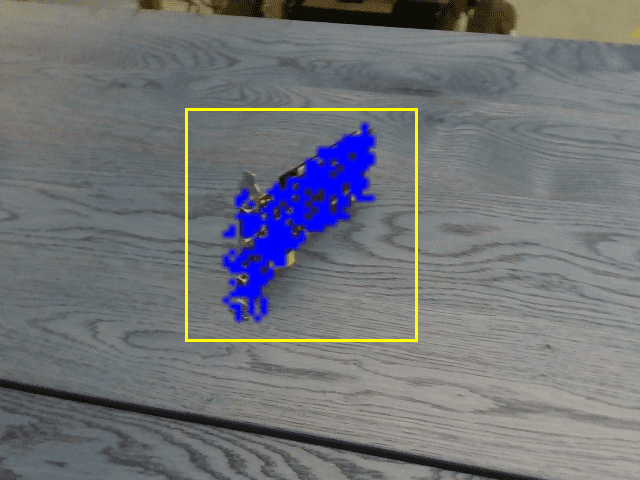} &
 \includegraphics[height=\qualmopoedheight]{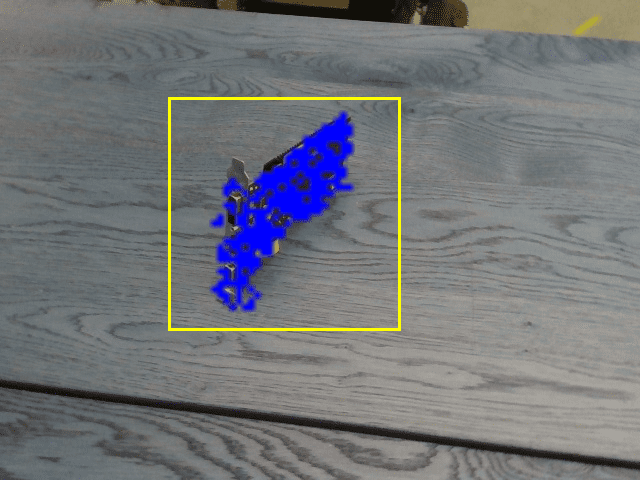} &
 \includegraphics[height=\qualmopoedheight]{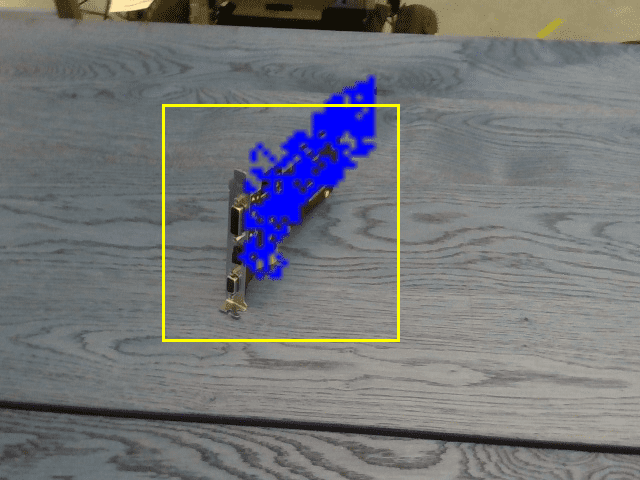}&
 \includegraphics[height=\qualmopoedheight]{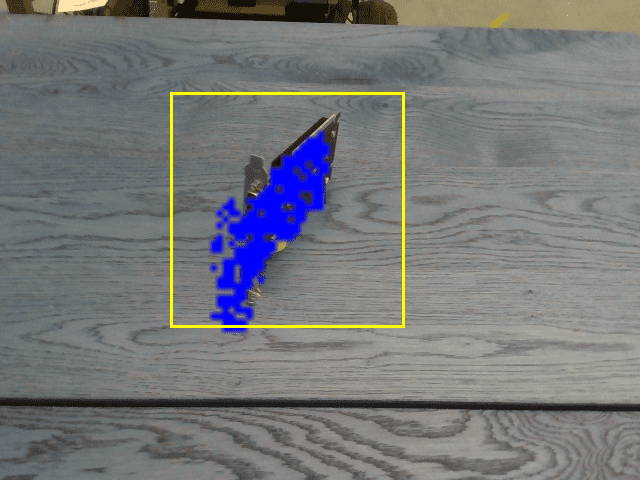}
\end{tabular}\\
\begin{tabular}{ccccc}
 \includegraphics[height=\qualmopoedheight]{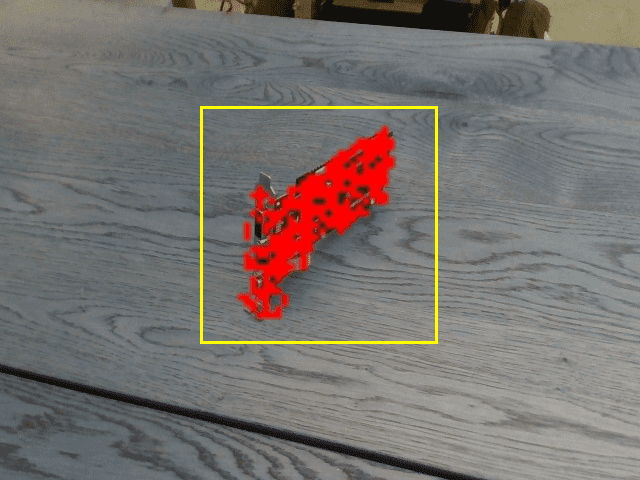} &
 \includegraphics[height=\qualmopoedheight]{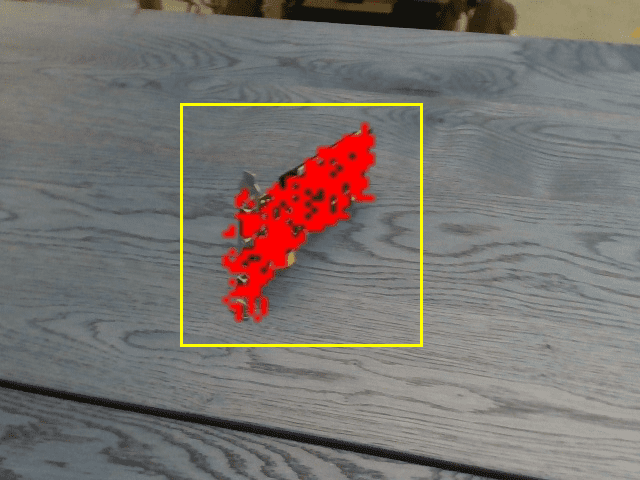} &
 \includegraphics[height=\qualmopoedheight]{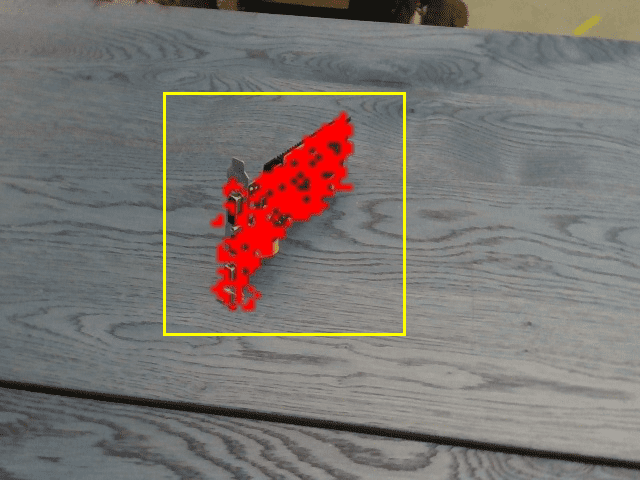} &
 \includegraphics[height=\qualmopoedheight]{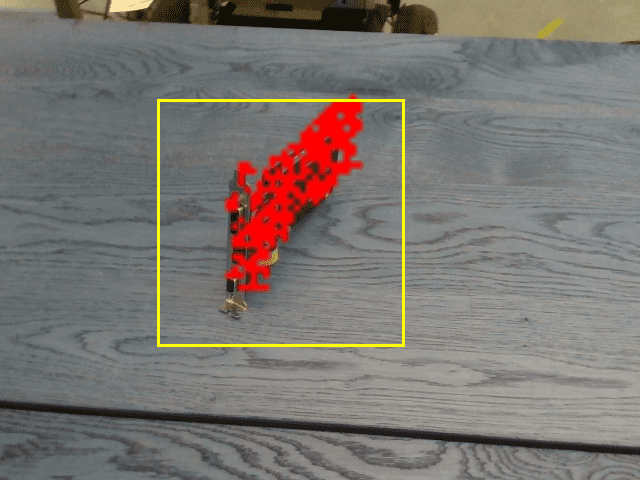}&
 \includegraphics[height=\qualmopoedheight]{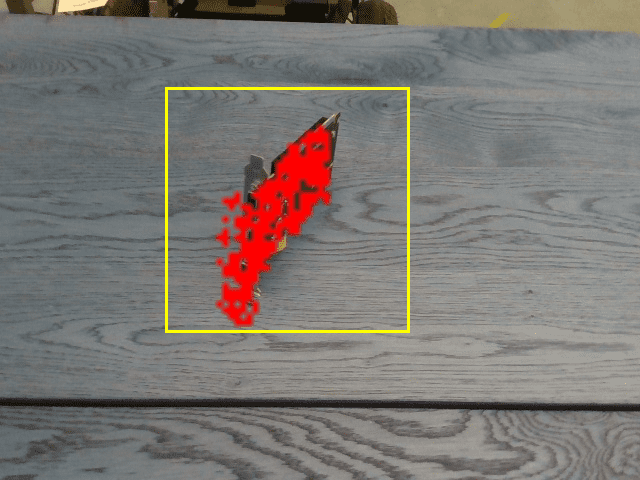}
\end{tabular}\\
\begin{tabular}{@{\;\;\;\;\;\;\;\;\;\;\;}c}
\vspace{3mm}
\includegraphics[height=0.20\qualmopoedheight]{images/introduction/time_arrow.pdf}
\end{tabular}
\end{center}
\vspace{-10mm}
\caption{\textbf{Qualitative results on real-world datasets.} We show in each image the ground-truth in \textbf{\color{green}green}, the prediction with Ours two-frame in \textbf{\color{blue}blue} and the prediction with Ours multi-frame in \textbf{\color{red}red}. Two first rows show our results on two sequences of object “clock" and “kinect" of Laval dataset \cite{garon2018framework}. The six last rows show our results on two sequences of object “rinse\_aid" and “graphic\_card" of Moped dataset \cite{park2020latent}.}
\label{fig:qualitative_results_real}

\end{figure*}

\noindent We show additional qualitative results on Moped~\cite{park2020latent} and Laval~\cite{garon2018framework} datasets in Figure~\ref{fig:segmentation_results} and Figure~\ref{fig:qualitative_results_real}. 


\end{document}